\documentclass[10pt,twocolumn,letterpaper]{article}

\usepackage{cvpr}            

\usepackage{graphicx}
\usepackage{amsmath}
\usepackage{amssymb}
\usepackage{booktabs}
\usepackage[dvipsnames]{xcolor}

\usepackage[pagebackref,breaklinks,colorlinks, allcolors=blue]{hyperref}

\usepackage[capitalize]{cleveref}
\crefname{section}{Sec.}{Secs.}
\Crefname{section}{Section}{Sections}
\Crefname{table}{Table}{Tables}
\crefname{table}{Tab.}{Tabs.}

\usepackage[accsupp]{axessibility}

\newcommand{\Mark}[1]{\textsuperscript{#1}}

\usepackage[utf8]{inputenc} 
\usepackage[T1]{fontenc}    
\usepackage{hyperref}       
\usepackage{url}            
\usepackage{booktabs}       
\usepackage{amsfonts}       
\usepackage{nicefrac}       
\usepackage{microtype}      
\usepackage{xcolor}         
\usepackage{multirow}
\usepackage{makecell}
\usepackage{graphicx}
\usepackage{comment}
\usepackage{bbding}
\usepackage{pifont}
\usepackage{subcaption}
\usepackage{listings}
\newcommand{\kETAL}    {{\em et al.}}

\begin{document}
\title{Towards Surveillance Video-and-Language Understanding: New Dataset, Baselines, and Challenges}

\author{Tongtong Yuan\Mark{1},
Xuange Zhang\Mark{1},
Kun Liu,
Bo Liu\Mark{1} \Mark{\Envelope}, 
Chen Chen\Mark{2}, 
Jian Jin\Mark{3},
Zhenzhen Jiao\Mark{4}
\\
\Mark{1}Beijing University of Technology, CN
\\
\Mark{2}Center for Research in Computer Vision, University of Central Florida, USA
\\
\Mark{3}Institute of Industrial Internet of Things, CAICT, CN
\\
\Mark{4}Beijing Teleinfo Technology Co., Ltd., CAICT, CN
\\
{\tt\small \{yuantt, liubo\}@bjut.edu.cn, chen.chen@crcv.ucf.edu}
}
\maketitle

\begin{abstract}
Surveillance videos are an essential component of daily life with various critical applications, particularly in public security. However, current surveillance video tasks mainly focus on classifying and localizing anomalous events. Existing methods are limited to detecting and classifying the predefined events with unsatisfactory semantic understanding, although they have obtained considerable performance.  
To address this issue, we propose a new research direction of surveillance video-and-language understanding, and construct the first multimodal surveillance video dataset. We manually annotate the real-world surveillance dataset UCF-Crime with fine-grained event content and timing. Our newly annotated dataset, UCA (\textbf{U}CF-\textbf{C}rime \textbf{A}nnotation), contains 23,542 sentences, with an average length of 20 words, and its annotated videos are as long as 110.7 hours. 
Furthermore, we benchmark SOTA models for four multimodal tasks on this newly created dataset, which serve as new baselines for surveillance video-and-language understanding. Through our experiments, we find that mainstream models used in previously publicly available datasets perform poorly on surveillance video, which demonstrates the new challenges in surveillance video-and-language understanding. To validate the effectiveness of our UCA, we conducted experiments on multimodal anomaly detection. The results demonstrate that our multimodal surveillance learning can improve the performance of conventional anomaly detection tasks. All the experiments highlight the necessity of constructing this dataset to advance surveillance AI. The link to our dataset is provided at: \url{https://xuange923.github.io/Surveillance-Video-Understanding}.
\end{abstract}    
\section{Introduction}

Surveillance videos are crucial and indispensable for public security. In recent years, various surveillance-video-oriented tasks have been widely studied, \eg, anomaly detection, anomalous/human action recognition, \etc. However, the existing surveillance video datasets~\cite{li2013anomaly,lu2013abnormal,Luo_2017_ICCV,sultani2018real} just provide the category labels and timing of anomalous events, and require all categories to be predefined. 
Thus, the related methods are still limited to detecting and classifying predefined events merely, lacking the deep semantic understanding capacity of video content. However, automatic understanding of surveillance video content is crucial to enhance the existing investigative measures in surveillance applications.
\begin{figure}
    \centering
    \includegraphics[width=1.\linewidth]{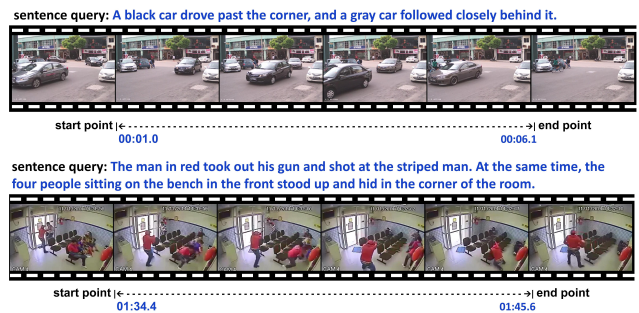}
    \caption{Annotation examples in our UCA dataset, including fine-grained sentence queries and the corresponding timing.}
    \vspace{-1em}
    \label{fig:datasample}
\end{figure}

In recent years, a large number of multimodal video datasets ~\cite{regneri2013grounding,rohrbach2015dataset,sigurdsson2016hollywood} have been released, on which various multimodal learning tasks~\cite{lan2023survey,abdar2023review,krishna2017dense} are being explored to semantically understand the video content. 
However, to our knowledge, surveillance-video-oriented multimodal learning is still understudied. For example, some tasks for investigating detailed cases in public security such as multimodal retrieval, and surveillance video captioning, have been rarely proposed and studied. 
One main reason is that the current surveillance datasets lack sentence-level language annotations, which hinders the learning of multimodal tasks. 
Another implicit reason is the increased difficulty in learning multimodal patterns from surveillance videos due to the unique characteristics they possess, which differentiate them from conventional video datasets. Therefore, this calls for a timely multimodal surveillance video dataset to validate the challenges of multimodal surveillance video learning and facilitate the development of surveillance AI.

To address these above issues, we propose extending the existing surveillance video datasets for anomaly detection to multimodal scenarios. 
Specifically, the multimodal surveillance video dataset should be composed of real-world videos, detailed event and timing descriptions, and as many labeled events as possible. Therefore, we investigated several surveillance datasets~\cite{adam2008robust,li2013anomaly,lu2013abnormal,Luo_2017_ICCV,sultani2018real,Cao_2023_CVPR} and selected UCF-Crime~\cite{sultani2018real} as the foundation of our new dataset. Because UCF-Crime is the largest real-world surveillance dataset and contains a variety of realistic anomalies. To obtain a multimodal surveillance dataset, we contribute new annotations for UCF-Crime~\cite{sultani2018real}. Our dataset is termed UCA (\textbf{U}CF-\textbf{C}rime \textbf{A}nnotation), and it is collected by making manually fine-grained annotations of event content and event timing on UCF-Crime~\cite{sultani2018real}. Some examples are shown in Figure~\ref{fig:datasample}. 

Compared with UCF-Crime, the main features of our UCA include: (1) The annotation information is relatively fine-grained, and includes as many event descriptions as possible. (2) Considering the realistic demand for temporal localization in the security field, our UCA also includes event timing along with the activity descriptions. (3) UCA can support multiple multimodal understanding tasks, such as Temporal Sentence Grounding in Videos (TSGV)~\cite{lan2023survey}, Video Captioning (VC)~\cite{abdar2023review}, Dense Video Captioning (DVC)~\cite{krishna2017dense}, Multimodal Anomaly Detection (MAD)~\cite{chen2023tevad}. Our main contributions can be summarized as follows: 
\begin{itemize}

    \item We establish the first real-world and multimodal surveillance dataset with both language descriptions and event timing, which is a comprehensive dataset for the surveillance field to develop the multimodal understandable capacity of machine intelligence.
    \item We are the first to establish the comprehensive baselines for multimodal tasks (\ie, TSGV, VC, and DVC) on surveillance videos. This research provides a foundational basis for understanding surveillance videos through the integration of video and language.
    \item We also demonstrate the effectiveness of our UCA in improving the existing surveillance applications. In experiments on multimodal anomaly detection (\ie, MAD), our methods can improve the performance of anomaly detection by providing a basic surveillance video captioning model as a plug-and-play module. 
    \item We thoroughly analyze these experiments and have discovered that mainstream models, commonly used in publicly conventional video-and-language understanding datasets, exhibit poor performance in multimodal surveillance video scenarios. This suggests that the basic learning model architectures must be modified and designed based on the unique characteristics of surveillance video datasets. This finding also underscores the necessity of constructing our dataset and highlights the challenges associated with multimodal surveillance video learning.
\end{itemize}

\section{Related Work}

\begin{table}[ht]
\caption{Comparison of the statistics of our UCA and other multimodal video datasets. \textbf{Avg word} means the average number of words per sentence. \textbf{Temp.anno.} means temporal annotation. The statistics of previous datasets have been recorded in~\cite{chen2019deep}.}
\centering
\resizebox{1.\linewidth}{!}{

\label{data comparison}
\begin{tabular}{lccccccc}
\toprule
Dataset & Domain &\#Videos & Duration(h) & \#Queries & Avg word  & Temp. anno. \\ \midrule
MSVD~\cite{chen2011collecting} & Open &1,970& 5.3 &70,028& 8.7  & \checkmark \\ 
TACos~\cite{regneri2013grounding} &Cooking &127 &10.1 &18,818 &9.0   &\checkmark  \\ 
YouCook~\cite{das2013thousand} &  Cooking &88 &2.3 &3,502 &12.6 &   \ding{55} \\ 
MPII-MD~\cite{rohrbach2015dataset} & Movie &94 & 73.6& 68,375& 9.6  &\checkmark \\ 
\thead{ActivityNet \\ Captions}~\cite{krishna2017dense} & Open& 20,000 & 849.0 &73,000 &13.5 & \checkmark \\ 
DiDeMo~\cite{anne2017localizing} & Open& 10,464 & 144.2 & 41,206 &7.5 & \checkmark \\ 
VATEX~\cite{Wang2019VaTeXAL} & Open& 41,250 & 114.6 & 825,000 &15.2 & \checkmark \\ 
MSRVTT~\cite{xu2016msr} & Open& 7,180 & 41.2 & 200,000 &9.3 & \checkmark \\ 
\midrule
\textbf{UCA (Ours)} &Surveillance & 1,854 & 121.9 &23,542 & 20.15 & \checkmark \\ 
\bottomrule
\end{tabular}
}
\end{table}

\subsection{Surveillance Video Datasets}

The majority of surveillance video datasets have some limitations on the number of videos or the degree of reality, such as UCSD Ped1~\cite{li2013anomaly}, UCSD Ped2 datasets~\cite{li2013anomaly}, Avenue dataset~\cite{lu2013abnormal}, Subway dataset~\cite{adam2008robust}, ShanghaiTech Campus dataset~\cite{Luo_2017_ICCV}, NWPU~\cite{Cao_2023_CVPR}, \etc. Differently, Sultani et al~\cite{sultani2018real} constructed a real-world surveillance video dataset, called UCF-Crime. The dataset consists of 1,900 surveillance videos that present 13 real-world anomalies, such as Abuse, Burglary, Explosion, etc. However, the annotation information of UCF-Crime only includes abnormal categories, which can only be used for abnormal detection. Thus, in the surveillance video research field, more complex multimodal learning tasks, such as moment retrieval and video captioning lack available datasets. 

To address this issue, we contribute new annotations on the largest surveillance video dataset UCF-Crime. We manually annotated the event content and event occurrence time for 1,854 videos from UCF-Crime, called \textbf{U}CF-\textbf{C}rime \textbf{A}nnotation (UCA). 
Additionally, we have noted that the paper SAVCHOI~\cite{mittal2022savchoi} mentions annotations on the UCF-Crime dataset for monitoring suspicious activities. However, SAVCHOI, which only annotated summaries for 300 videos and lacked detailed temporal information on activities, is not directly comparable with our UCA. The annotations from SAVCHOI are not included in our UCA dataset.


\subsection{Multimodal Video Datasets}

Recently, numerous video datasets have been released for different video-language-understanding tasks, such as video caption, video dense caption, temporal sentence grounding in videos (TSGV), \etc. We review some video datasets widely utilized in various video-understanding tasks, and list the comparison of the statistics of our UCA and other seven conventional video datasets for multimodal learning tasks in Table~\ref{data comparison}. All of these video datasets consist of high-quality videos and have been annotated with detailed descriptions, which have played indispensable roles in video and language tasks. The main difference between our dataset and other multimodal datasets lies in the difference in domain. Our dataset is specifically designed for the surveillance field. Additionally, there are also differences in aspects such as video numbers, duration, queries, \etc. It is worth noting that our UCA has the longest word length.

\subsection{Multimodal Video Learning}

In the following, we briefly review some mainstream multimodal video learning tasks. For more completed reviews, we refer readers to~\cite{liu2023survey,li2019visual,chen2019deep,abdar2023review,csengonul2023analysis}.

\begin{itemize}
    
\item Temporal sentence grounding in videos (TSGV) is a recent multimodal task~\cite{liu2023survey,lan2023survey,zhang2022temporal}, which learns the temporal activity localization from a given video with respect to a given language query, \ie, the goal of the task is localizing the start and end times for the described activity inside the video. TSGV can be applied in surveillance video to retrieve the event and moment.


\item Video captioning (VC) is a multimodal task of producing a natural-language utterance, which aims at describing the visual content of a video. It plays an important technical role in the demand for automatic visual understanding and content summarization~\cite{li2019visual,abdar2023review}, which is also important in surveillance video understanding. 


\item Dense video captioning (DVC) aims to acquire the temporal localization and captioning of all events in an untrimmed video. Surveillance video analysis can also benefit from Dense video captioning. It is more challenging than standard video captioning, which aims to generate a single caption for a video clip~\cite{abdar2023review,krishna2017dense}. 
\item Multimodal anomaly detection (MAD) is proposed by Chen~\kETAL ~\cite{chen2023tevad}, which fused video caption features with original visual and temporal features for anomaly detection in surveillance videos. But the video caption information is derived from Swinbert~\cite{lin2022swinbert} trained on open-domain dataset VATEX~\cite{Wang2019VaTeXAL}. We aim to introduce surveillance-domain VC models to generate captions to enhance the multimodal anomaly detection task.
\end{itemize}
With the manually fine-grained annotations, our UCA dataset can be used in various multimodal video learning tasks, including but not limited to the above four tasks. We select SOTA methods of these tasks as baselines in the following experiments, by considering various factors, including the novelty, the differences between methods, the stability of experimental results, and the openness of the code, \etc. These baselines are detailed in \textbf{Sec. 4}. 
\section{The UCA Dataset}
Our dataset is based on the UCF-Crime dataset, which is a real-world surveillance video dataset containing 13 real-world anomalies and some normal videos. To extend UCF-Crime to a multimodal dataset, we conducted a fine-grained language annotation on UCF-Crime that recorded each event/change of the videos with detailed descriptions and time stamps. The result of our annotation is a novel dataset named UCA, which is the first large-scale multimodal surveillance video dataset for TSGV, VC, DVC, and MAD.

In this section, we provide a comprehensive outline of the UCA dataset, covering the aspects of data collection and annotation, dataset analysis, comparison with existing datasets, application scope, and ethical considerations. By presenting these details, we aim to provide a concise understanding of the UCA dataset and its potential for advancing research in the field of surveillance video-and-language understanding.

\begin{table}[t]
\caption{Data split and video statistics of our UCA dataset.}
\centering
\label{Data1}
\resizebox{1.\linewidth}{!}{
\begin{tabular}{cccccc}
\toprule
Video Statistics &
  {UCF-Crime} &
  {UCA} &
  {UCA train} &
  {UCA val} &
  {UCA test} \\ \midrule
\#Video       & 1,900 & 1,854   & 1,165 & 379   & 310   \\ 
Video length           & 127.5h & 121.9h   & 75.5h  & 21.2h& 25.2h  \\ 
Annotated video length & —    & 110.7h & 73.7h& 16.4h  & 20.6h  \\ 
\bottomrule
\end{tabular}%
}
\end{table}

\subsection{Collection and Annotation}

{During the video collection, we filtered some low-quality videos in the original UCF-Crime to ensure the quality and fairness of our UCA. These low-quality videos encompassed instances of repetition, severe occlusion, and excessively accelerated playback, which affected the clarity of manual annotations and the precision of event timing localization. Therefore, we removed 46 videos from the original UCF-Crime dataset, resulting in 1,854 videos for UCA. The data split in UCA is shown in Table~\ref{Data1}. 

When labeling a video from UCF-Crime, our goal is to make fine-grained annotations, \ie, making a detailed description of each event that can be described in language as much as we can, regardless of whether it is an abnormal event. We also record the starting and ending time for each event in an individual video and the recorded time is 0.1-second interval. The annotated video length of UCA is 110.7 hours, accounting for 86.8\% of the total video duration of UCA. Figure \ref{fig:datasample} presents some annotation examples in UCA. 

During the dataset annotation process, we recruited 10 volunteers with computer backgrounds as annotators and formed a review team consisting of 3 AI researchers. To ensure the accuracy and consistency of the annotations, we provided the annotators with comprehensive annotation instructions (The instructions are shown in \textbf{Sec. A} of Appendix). These guidelines were designed to ensure that the annotators utilize linguistically informative language when describing the events depicted in the videos, thereby ensuring clarity and accuracy. Additionally, the instructions emphasized the accurate recording of the start and end times.
\\
Before commencing the annotation work, we provided all annotators with comprehensive training. Throughout the annotation process, annotators were required to watch the videos repeatedly to ensure accurate positioning and description of events. In order to maintain the annotation quality, we implemented a validation process conducted by the review team, wherein annotators' work was reviewed every 100 instances. The focus of this validation was on the quality and consistency of annotations provided by different annotators. Following the completion of all annotations (23,542 sentence-level queries), reviewers conducted a further review of annotated data. The entire annotation and review process required approximately two months. A detailed description of the annotation procedure is shown in Appendix Sec. A.

\begin{figure*}[htb]
    \centering
    \includegraphics[width=0.98\linewidth]{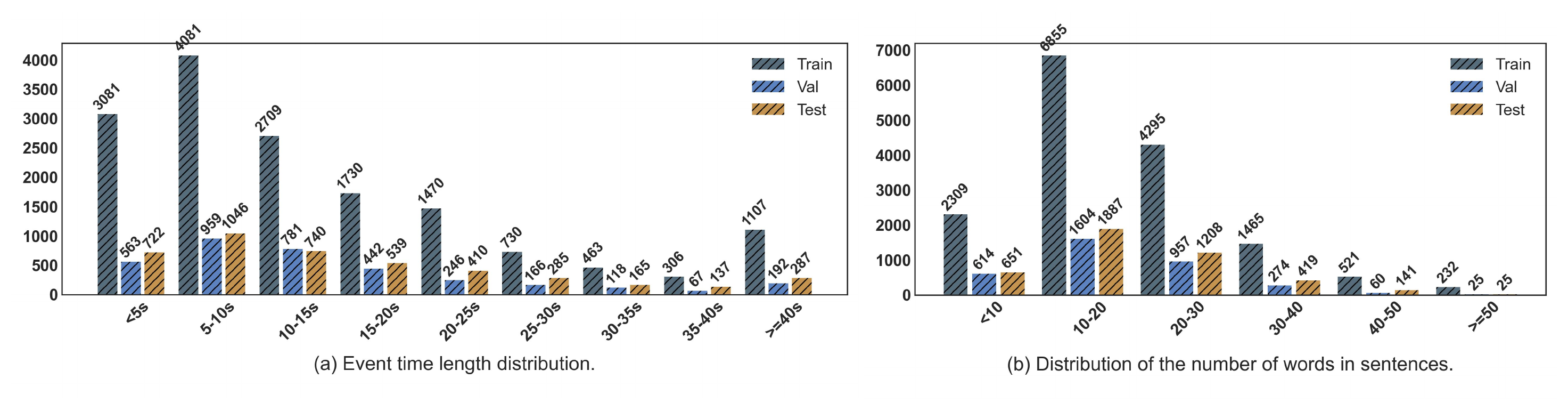}
    \vspace{-1em}
    \caption{The duration of annotated events and the number of words in the annotated queries of the UCA dataset.}
    \label{fig:merge}
\end{figure*}

\subsection{Dataset Analysis}

As shown in Table \ref{Data1}, the UCA dataset comprises a total of 1854 videos, which is divided into three subsets: Train, Validation, and Test sets according to the video lengths and original video categories.
These abnormal instances encompass a spectrum of 13 real-world anomalies, including \textit{Abuse, Arrest, Arson, Assault, Burglary, Explosion, Fighting, RoadAccidents, Robbery, Shooting, Shoplifting, Stealing, and Vandalism}~\cite{sultani2018real}. Details for video numbers of different categories in UCA are shown in Appendix A. UCA has little difference from the original UCF-Crime dataset in the modality of video. Therefore, we focus on analyzing statistical information on the language modality.

It can be seen that the total duration of the annotated video is 110.7 hours as shown in Table~\ref{Data1}. Figure~\ref{fig:merge} (a) shows the distribution of the time length of each annotated event in our dataset. The distribution indicates that the video length of the event with the most number is 5-10 seconds, with the majority of events concentrated within 30 seconds and a few events exceeding 40 seconds.

Table~\ref{data3} shows the number of query descriptions of the events we labeled and the average number of words per query. It can be seen that the average number of words in our annotation is around 20 words. The distribution of the number of words in annotated queries of UCA is shown in Figure~\ref{fig:merge} (b). The highest number of sentences contains 10-20 words, followed by sentences containing 20-30 words. Additionally, we perform the vocabulary statistics in UCA queries, which can be referred to Table~\ref{data3}. We can find the ratio of nouns, verbs, and adjectives in all sentences of Train, Val, and Test is approximately 2:2:1. This indicates a consistent annotation distribution across different videos.

\begin{table}[h]
\caption{Statistics of annotation in our UCA dataset, including the number of queries (\#Query), the number of words per query (\#Word/Query), as well as the numbers of nouns (\#Nouns ), verbs (\#Verbs), and adjectives (\#Adj).}
\vspace{-1em}
\centering
\label{data3}
\resizebox{0.85\linewidth}{!}{
\begin{tabular}{lcccc}
\toprule
Annotation Statistics & \#Train & \#Val & \#Test & \#Summary \\ \midrule
\#Query          & 15,677         & 3,534   & 4,331                    & 23,542            \\ 
\#Word/Query           & 20.45        & 18.82      & 20.13                & 20.15            \\ 
\#Nouns            & 25,333           & 5,633           & 6,748          & 37,714            \\ 
\#Verbs            & 26,817           & 5,569           & 6,959          & 39,345            \\ 
\#Adj              & 12,880           & 2,963           & 3,571          & 19,414      \\ \bottomrule
\end{tabular}
}
\end{table}

\subsection{Comparison with Existing Datasets}

\textbf{Difference with video-language datasets.}
Table~\ref{data comparison} presents a comparison between our UCA dataset and video-language datasets for multimodal learning tasks. Our UCA is for the surveillance domain, which distinguishes it from other datasets. Through this comparison, we observe that our dataset has a moderate number of annotated sentences. However, in terms of the average number of words per sentence, our dataset has the highest number of annotations, indicating that our annotated sentence descriptions are more specific than those of other datasets. \textbf{The main differences between UCA and other conventional video datasets lie in the domain of videos (the domain can be seen in Table~\ref{data comparison}) and video quality (such as uneven image quality, complex backgrounds, and complex events in surveillance videos).} Consequently, when faced with the same learning task, the learning difficulty on our dataset is often much higher than that on a conventional video dataset. The subsequent experiments will demonstrate the experimental results of some state-of-the-art learning methods tested on our dataset, indicating the challenges in multimodal surveillance video learning.

\textbf{Difference with abnormal video datasets.}
Our work highlights a key difference from abnormal video datasets~\cite{sultani2018real,Luo_2017_ICCV}: we conducted the unprecedented large-scale sentence-level annotation of events in real-world surveillance scenarios. In contrast to the simple category annotation, we provide specific event descriptions in a fine-grained way. As a result, our dataset can be used to conduct various video-and-language understanding tasks in surveillance scenarios. To our knowledge, research on multimodal surveillance learning is still blank, and it holds great practical significance.

\subsection{Application Scope}
UCA is the first multimodal surveillance video dataset and can be applied to the research on intelligent public security. In particular, it can provide a fundamental dataset for tasks related to multimodal surveillance video comprehension, which is a novel area within the surveillance domain. Based on UCA, researchers can explore retrieving detailed event queries with temporal information, captioning surveillance videos, and multimodal anomaly detection, to improve technical investigative capabilities. 

\subsection{Ethical Considerations}
The videos utilized in our research are sourced from a publicly available dataset called UCF-Crime~\cite{sultani2018real}. This dataset does not explicitly mention any privacy or ethical concerns. Our work exclusively involves sentence-level annotations to describe the events depicted in these videos. Throughout our annotation process, we discovered that the primary videos did not exhibit any particular focus on race or gender. To further minimize any potential risks related to data annotation, we implemented multiple measures, such as annotation guidance and review procedures. Our annotation examples are shown in Figure~\ref{fig:datasample}, and more examples are provided in Appendix Sec. A. Additionally, we also provide gender-neutral annotations by substituting terms like "man" and "woman" in our annotations with the term "person". Without loss of generality, the experiments presented in this paper were conducted without using gender-neutral annotations.

Users are expected to adhere to our data license agreement, which can be accessed in Appendix Sec. D. We emphasize that the usage of the dataset is strictly limited to academic and research purposes. We encourage the research community to provide suggestions for enhancing the quality and ethical reliability of the dataset.

\section{Experiments}
We proceed with the experiments of four multimodal tasks on the UCA dataset using an RTX3090 GPU. In each task, we first describe the corresponding task and evaluation metric, along with baseline methods, and then present the performance of our selected baselines. We followed the basic experimental settings of these various methods in their respective published papers and public code repositories. Our codebase including data processing and baselines will be open-source to encourage researchers to integrate their new models. For readability, more detailed experiment settings, complementary experiment results, and more visualization examples can be found in Appendix Sec. B and Sec. C.

\subsection{Temporal Sentence Grounding in Videos}

\textbf{Task.}
Temporal sentence grounding in videos (TSGV) aims to learn the temporal activity localization from a given video with respect to a given language query~\cite{liu2023survey}.

\textbf{Metric.}
In existing methods~\cite{gao2017tall,yuan2019semantic,zhang2021multi}, R@$K$ for IoU=$\theta$ is commonly adopted as the evaluation metric to measure the performance in TSGV. It is defined as the percentage of at least one of the top-$K$ predicted moments that have
IoU with ground-truth moment larger than $\theta$~\cite{gao2017tall}. In the following, we set R@$K$ for IoU$=\theta$ with $K={1,5}$ and $\theta={0.3,0.5,0.7}$ as the evaluation metric.

\begin{table}[t]
\centering
\caption{Benchmarking of TSGV baselines on our UCA dataset.}
\label{tab:TSGV}
\resizebox{0.9\linewidth}{!}{%
\begin{tabular}{lcccccc}
\toprule
\multirow{2}{*}{Method}  & \multicolumn{2}{c}{IoU=0.3}      & \multicolumn{2}{c}{IoU=0.5}      & \multicolumn{2}{c}{IoU=0.7}      \\ \cmidrule{2-7}
                                                                                     & R@1 & R@5   & R@1  & R@5   &R@1  & R@5    \\ \midrule
CTRL~\cite{gao2017tall}                                                                             & \multicolumn{1}{c}{6.35} & 20.09 & \multicolumn{1}{c}{3.00} & 10.88 & \multicolumn{1}{c}{0.74} & 3.32                           \\ 
SCDM~\cite{yuan2019semantic}                                                                               & \multicolumn{1}{c}{6.49} & 10.72 & \multicolumn{1}{c}{3.65} & 7.39  & \multicolumn{1}{c}{1.66} & 4.78                           \\ 
A2C~\cite{he2019read}                                                                               & \multicolumn{1}{c}{4.25} &   -    & \multicolumn{1}{c}{1.78} &    -   & \multicolumn{1}{c}{0.44} &   -                           \\ 
2D-TAN ~\cite{zhang2020learning}                                                    & \multicolumn{1}{c}{7.76} & 19.67 & \multicolumn{1}{c}{4.16} & 11.54 & \multicolumn{1}{c}{2.22} & 6.00                       \\ 
LGI ~\cite{mun2020local}                                                   & \multicolumn{1}{c}{7.71} & - & \multicolumn{1}{c}{3.26} & - & \multicolumn{1}{c}{1.18} & -                       \\ 
MMN ~\cite{wang2022negative}                                                            & \multicolumn{1}{c}{8.68} & 21.38 & \multicolumn{1}{c}{4.66} & 11.98 & \multicolumn{1}{c}{2.12} & 5.89        \\ 

MomentDiff ~\cite{li2023momentdiff}                                                          & \multicolumn{1}{c}{8.15} & - & \multicolumn{1}{c}{4.46} & - & \multicolumn{1}{c}{1.71} & -        \\ 

\bottomrule

\end{tabular}}%
\end{table}

\begin{figure}[ht]
    \centering
    \includegraphics[width=1.\linewidth]{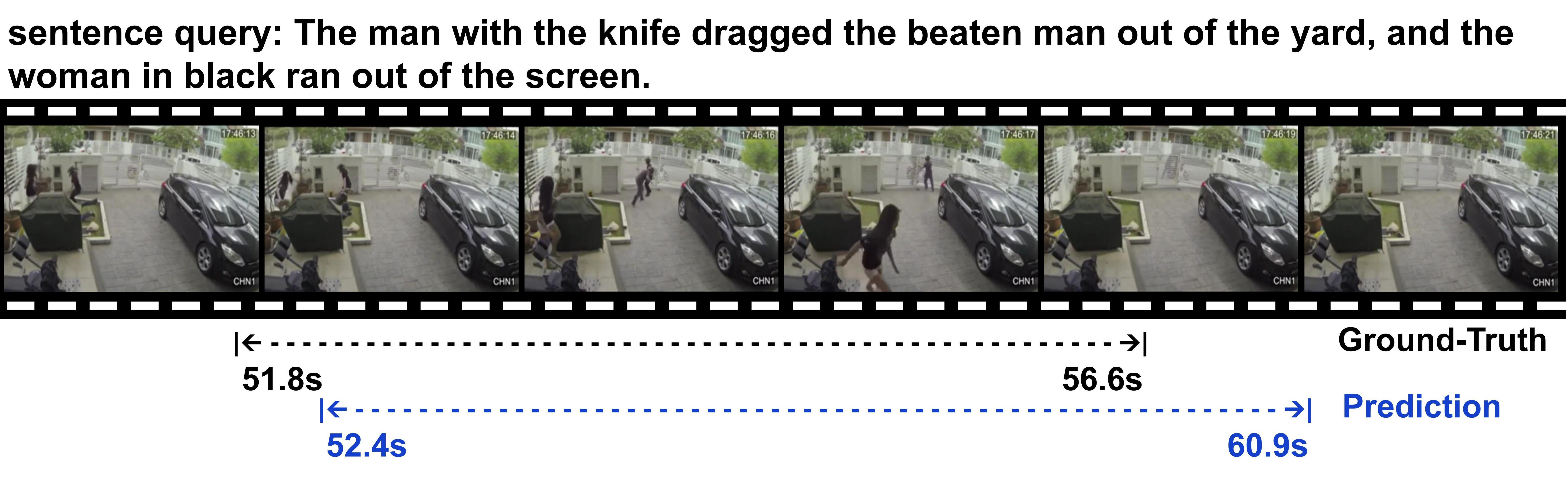}
    \caption{A visualization example of TSGV by MMN~\cite{wang2022negative}.}
    \vspace{-1em}
    \label{fig:TSGV}
\end{figure}

\textbf{Baselines.}
We benchmark UCA using 7 different methods (ranging from 2017 to 2023), comparing the novelty, differences, and reproducibility of experiments. CTRL~\cite{gao2017tall} represented a traditional sliding window-based framework of TSGV. SCDM~\cite{yuan2019semantic} introduced the Semantic Conditioned Dynamic Modulation mechanism, which is a standard anchor-based Method. A2C~\cite{he2019read} applied reinforcement learning to this field. For 2D-Map anchor-based methods, 2D-TAN~\cite{zhang2020learning} proposed a unique two-dimensional temporal matrix, while MMN~\cite{wang2022negative} further introduced the idea of metric learning. LGI~\cite{mun2020local} proposed a local-global interaction (LGI) framework as a regression-based method. Momentdiff~\cite{li2023momentdiff} applied a diffusion model to video moment retrieval.

\textbf{Implementation Settings.} We use C3D~\cite{tran2015learning} pretrained on Sports1M to extract video features for all experiments and the 4,096 dimensions output of the FC6 layer is represented as the video feature. For CTRL, we employ skip-thought~\cite{kiros2015skip} as the sentence encoder, resulting in sentence features of 4,800 dimensions. For SCDM, the length of input video clips is set as 512 to accommodate the temporal convolution. For A2C, we also obtain sentence features via skip-thought. For 2D-TAN, the number of video clips is set to 16, and the scaling thresholds $t_{min}$ and $t_{max}$ are set to 0.5 and 1.0. We chose the stacked convolution approach to extract moment-level features. For LGI, the number of sampled segments per video is 128, and the maximum length of each query is 50. The settings of MMN are similar to 2D-TAN, except that max-pooling is used when obtaining moment-level features. The text encoder of Momentdiff utilizes Glove, with a maximum text length set to 32.

\textbf{Results and Analysis.} The experimental results from Table~\ref{tab:TSGV} indicate that the TSGV task presents significant challenges within the domain of surveillance videos.  In the R@$1$, IoU = $0.3$ metrics, all evaluation results are generally below 10\%. This underscores the difficulty in achieving precise event localization in surveillance videos. The videos in UCA have various video durations, but existing models do not adequately consider the handling of long-term temporal information, thereby further affecting the comprehensive fusion of text and video feature information. Among the numerous benchmarking methods, the A2C model, which employs reinforcement learning mechanisms, exhibits relatively inferior performance on the UCA dataset. However, two models employing 2D Temporal Adjacent Networks, namely 2D-TAN and MMN, demonstrate superior overall performance compared to other methods. The visualization example of TSGV by MMN is shown in Figure~\ref{fig:TSGV}. This advantage can possibly be attributed to their capability to perceive richer contextual information from video data, thereby better comprehending and capturing these temporal characteristics.

\subsection{Video Captioning}

\begin{figure}[h]
    \centering
    \includegraphics[width=1.\linewidth]{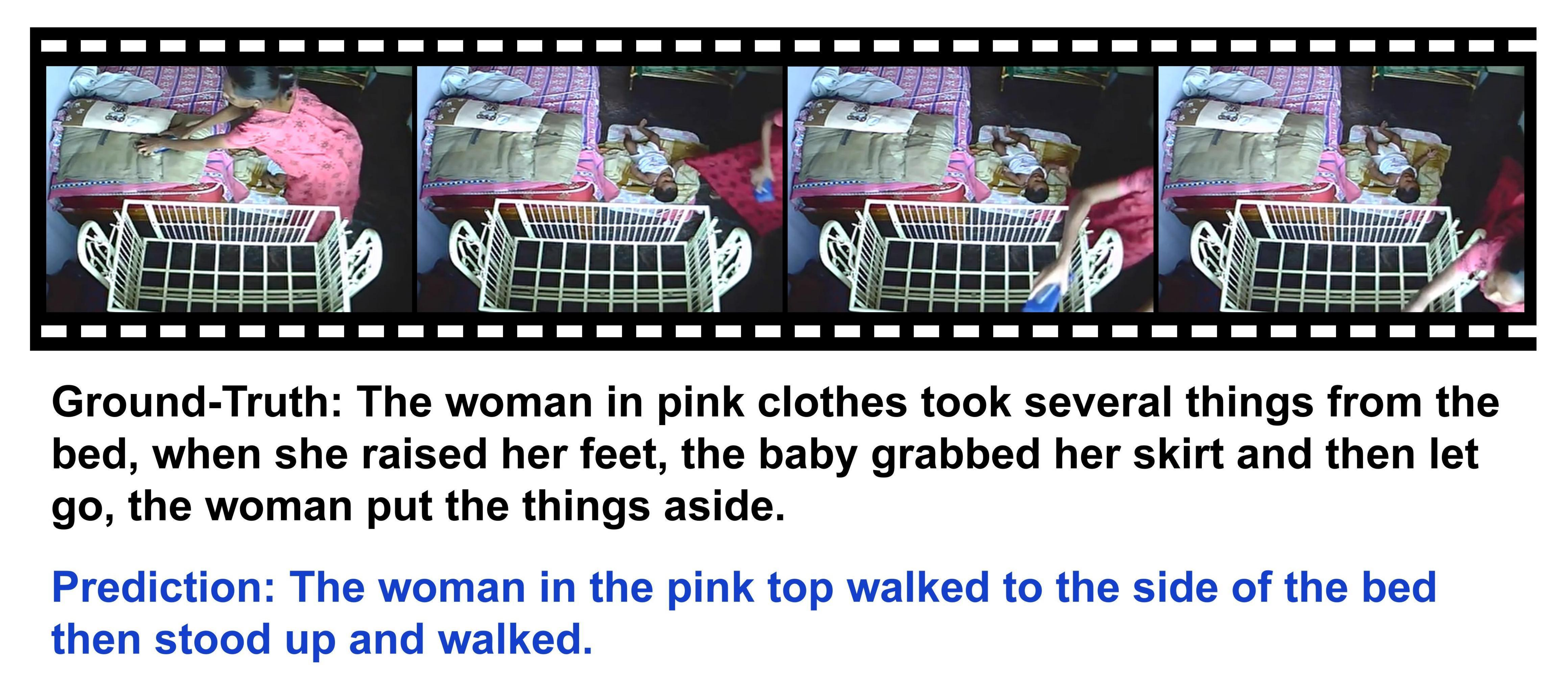}
    
    \caption{A visualization example of VC by SwinBert~\cite{lin2022swinbert}.}
    \label{fig:VC}
\end{figure}

\textbf{Task.} The goal of video captioning (VC) is understanding a video clip and describing it with language~\cite{abdar2023review}.

\textbf{Metric.}
We use the metrics as in ~\cite{lin2022swinbert,ryu2021semantic}. 
The evaluation metrics of correctness
include Bilingual Evaluation Understudy (BLEU) [B@n, n=1,2,3,4]~\cite{papineni2002bleu},
Metric for Evaluation of Translation with Explicit Ordering
(METEOR) [M]~\cite{denkowski2014meteor}, Recall Oriented Understudy of Gisting
Evaluation (ROUGE-L) [R]~\cite{lin2004rouge}, and Consensus-based Image Description Evaluation (CIDEr) [C] ~\cite{vedantam2015cider}.
\begin{table}[ht]
\centering
\caption{Benchmarking of VC baselines on our UCA. B, M, R, and C represent BLEU, METEOR, ROUGE-L, and CIDEr, respectively.}
\label{video}
\resizebox{1.\linewidth}{!}{%
\begin{tabular}{llccccccccc}
\toprule
Method         & Features                                             & B1    & B2    & B3   & B4   & M    & R     & C     \\ \midrule
S2VT~\cite{venugopalan2015sequence}           & Inception V4                                                            & 28.27 & 16.34 & 9.62 & 5.56 & 10.35 & 25.38 & 15.44         \\ 
S2VT~\cite{venugopalan2015sequence}          & VGG 16                                                                    & 23.53 & 13.17 & 7.72 & 4.53 & 10.39 & 24.09 & 13.92         \\ 
RecNet\_global~\cite{wang2018reconstruction}  & Inception V4                                                              & 25.81 & 15.06 & 9.14 & 5.78 & 10.42 & 25.12 & 16.90         \\ 
RecNet\_local~\cite{wang2018reconstruction}  & Inception V4                                                             & 26.89 & 15.73 & 9.52 & 6.00 & 10.55 & 25.52 & 16.70           \\
 
MARN~\cite{pei2019memory}          & \begin{tabular}[c]{@{}c@{}}Inception V4\\ ResNext101\end{tabular}          & 26.16 & 15.61 & 10.22 & 6.63 & 9.67 & 23.85 & 15.33          \\ 
SGN~\cite{ryu2021semantic}            & \begin{tabular}[c]{@{}c@{}}ResNet101\\ ResNext101\end{tabular}             & 29.17 & 16.93 & 10.32 & 6.28 & 11.73 & 26.25 & 18.95         \\
SwinBERT~\cite{lin2022swinbert}      & VidSwin                                                              & 25.02 & 15.60  & 9.98 & 6.33 & 11.15 & 27.16 & 25.29            \\
CoCap~\cite{shen2023accurate}      & CLIP                                                                &28.53  &17.34   &10.75  &6.57  &11.43&28.14  &21.24            \\
\bottomrule
\end{tabular}}%
\end{table}

\textbf{Baselines.} We select 6 methods (ranging from 2015 to 2023) for video captioning as our baselines. These methods have various motivations for learning VC: S2VT~\cite{venugopalan2015sequence} is a classic work that first proposed an end-to-end model for generating video descriptions, providing important insights for subsequent research. RecNet~\cite{wang2018reconstruction} introduced the encoder-decoder-reconstructor architecture, using the idea of forward and backward flows for video caption generation. MARN~\cite{pei2019memory} improved the traditional encoder-decoder framework and proposed a Memory-Attended Recurrent Network. SGN~\cite{ryu2021semantic} emphasized the importance of decoded captions and used semantic groups as information units to describe videos. SwinBERT~\cite{lin2022swinbert} is the first end-to-end video caption generation model fully based on Transformer. CoCap~\cite{shen2023accurate} introduced an end-to-end video captioning approach based on compressed videos, achieving faster runtime speeds.

\textbf{Implementation Settings.} The video feature extractor settings are shown in Table~\ref{video}. For SwinBert, the VidSwin is initialized with Kinetics-600 pre-trained weights~\cite{liu2022video}. For SGN, we uniformly sample 50 frames and clips from each video, and the visual encoder utilizes pre-trained ResNet and 3D-ResNext~\cite{hara2018can} models. ResNext in MARN and SGN corresponds to the motion feature, which focuses on the dynamic variation information between video frames. 

\textbf{Results and Analysis.} The performance of video captioning baselines is presented in Table~\ref{video}. Among these benchmarks, SGN has achieved a higher accuracy in phrase matching due to its semantic group design, effectively integrating annotation information with video frames. Two Transformer-based models, SwinBERT and CoCap, have demonstrated superior performance across multiple metrics, particularly in this CIDEr metric. A visualization example of Swinbert is shown in~\ref{fig:VC}. This substantiates their capability to capture key textual information more effectively. Overall, surveillance videos have distinctive characteristics, such as blurry and low-resolution visuals, which will pose more challenges for VC models to learn and align the visual modality and the text modality.

\subsection{Dense Video Captioning}
\vspace{-1em}
\begin{table}[ht]
    \centering
    \caption{Event localization of DVC on the UCA dataset.}
    \vspace{-1em}
    \label{dense event}
\resizebox{1.\linewidth}{!}{

\begin{tabular}{c c cccc cccc}
\hline
\multirow{2}{*}{Method} & \multirow{2}{*}{Features} & \multicolumn{4}{c}{Recall}                                                                                               & \multicolumn{4}{c}{Precision}                                                                                          \\ \cline{3-10} 
                        &                           & \multicolumn{1}{c}{0.3}   & \multicolumn{1}{c}{0.5}   & \multicolumn{1}{c}{0.7}   & \multicolumn{1}{c}{0.9}      & \multicolumn{1}{c}{0.3}   & \multicolumn{1}{c}{0.5}   & \multicolumn{1}{c}{0.7}  & \multicolumn{1}{c}{0.9}   \\ \hline
TDA-CG~\cite{wang2018bidirectional}                 & C3D                       & \multicolumn{1}{c}{32.54} & \multicolumn{1}{c}{16.64} & \multicolumn{1}{c}{8.40} & \multicolumn{1}{c}{2.49}  & \multicolumn{1}{c}{58.79} & \multicolumn{1}{c}{27.16} & \multicolumn{1}{c}{8.95} & \multicolumn{1}{c}{1.44}  \\ 
PDVC~\cite{wang2021end}                  & C3D                       & \multicolumn{1}{c}{48.03} & \multicolumn{1}{c}{28.48} & \multicolumn{1}{c}{11.80} & \multicolumn{1}{c}{2.50} & \multicolumn{1}{c}{67.33} & \multicolumn{1}{c}{36.19} & \multicolumn{1}{c}{12.20} & \multicolumn{1}{c}{1.94}  \\ 
TDA-CG~\cite{wang2018bidirectional}                 & I3D                       & \multicolumn{1}{c}{31.72} & \multicolumn{1}{c}{17.09} & \multicolumn{1}{c}{8.97} & \multicolumn{1}{c}{2.89}  & \multicolumn{1}{c}{57.88} & \multicolumn{1}{c}{29.69} & \multicolumn{1}{c}{12.45} & \multicolumn{1}{c}{1.91}  \\ 
PDVC~\cite{wang2021end}                   & I3D                       & \multicolumn{1}{c}{49.53} & \multicolumn{1}{c}{28.59} & \multicolumn{1}{c}{12.56} & \multicolumn{1}{c}{2.11}  & \multicolumn{1}{c}{68.84} & \multicolumn{1}{c}{35.36} & \multicolumn{1}{c}{12.84} & \multicolumn{1}{c}{1.73}  \\ \hline
\end{tabular}
  }
\end{table}
\vspace{-1em}
\begin{table}[h]
    \centering
    \caption{DVC on the UCA dataset with predicted proposals.}
    \vspace{-1em}
    \label{dense predicted}
\resizebox{1.\linewidth}{!}{
\begin{tabular}{ccccccccc}
\hline
\multicolumn{1}{c}{\multirow{2}{*}{Method}} & \multicolumn{1}{c}{\multirow{2}{*}{Features}} & \multicolumn{7}{c}{Predicted proposals}                                                                                                                                                            \\ \cline{3-9} 
\multicolumn{1}{c}{}                        & \multicolumn{1}{c}{}                          & \multicolumn{1}{c}{B1}   & \multicolumn{1}{c}{B2}   & \multicolumn{1}{c}{B3}   & \multicolumn{1}{c}{B4}   & \multicolumn{1}{c}{M}    & \multicolumn{1}{c}{C}    & \multicolumn{1}{l}{SODA\_c} \\ \hline
TDA-CG~\cite{wang2018bidirectional}                                     & C3D                                            & 5.50                     & 2.23                      & 0.84                      & 0.38                      & 2.72                      & 4.44                      & 0.98                        \\
PDVC~\cite{wang2021end}                                         & C3D                                            & 7.89                      & 3.93                      & 1.65                      & 0.55                      & 3.78                      & 8.58                      & 2.07                        \\
TDA-CG~\cite{wang2018bidirectional}                                         & I3D                                            & 6.35                      & 2.84                      & 1.22                      & 0.37                      & 3.12                      & 5.51                      & 0.98                        \\ 
\multicolumn{1}{c}{PDVC~\cite{wang2021end}}                   & \multicolumn{1}{c}{I3D}                       & \multicolumn{1}{c}{8.02} & \multicolumn{1}{c}{4.22} & \multicolumn{1}{c}{2.00} & \multicolumn{1}{c}{0.71} & \multicolumn{1}{c}{4.06} & \multicolumn{1}{c}{8.78} & \multicolumn{1}{c}{2.21}   \\ \hline
\end{tabular}
  }
\end{table}

\textbf{Task.}
Dense video captioning (DVC) learns to generate the temporal localization and captioning of dense events in an untrimmed video. 

\textbf{Metric.}
We have assessed the performance from two perspectives. For localization performance, we employed the evaluation tools provided by the ActivityNet Captions Challenge 2018 and used the common metrics~\cite{wang2018bidirectional,zhang2022unifying}, like different IoU thresholds (0.3, 0.5, 0.7, 0.9), classic caption evaluation metrics: BLEU, METEOR, CIDEr, and the performance in describing video stories: SODA\_c score.

\textbf{Baselines.}
We select three methods (ranging from 2018 to 2022) for DVC with different input features (to increase the diversity of comparisons) as our baselines. These methods have distinct designs and are reproducible. 
TDA-CG~\cite{wang2018bidirectional} proposed a bidirectional proposal method that effectively utilizes past and future context for proposal prediction. PDVC~\cite{wang2021end} enhanced the representation of event queries and inputs the localization head and caption head in parallel. UEDVC~\cite{zhang2022unifying} fully utilized the interdependence between events to detect more diverse and consistent events.

\begin{table}[htb]
    \centering
    \caption{DVC on the UCA dataset with Ground-Truth proposals.}
    \vspace{-1em}
    \label{dense ground}
\resizebox{1.\linewidth}{!}{
\begin{tabular}{cccccccc}
\hline
\multicolumn{1}{c}{\multirow{2}{*}{Method}} & \multicolumn{1}{c}{\multirow{2}{*}{Features}} & \multicolumn{6}{c}{Ground-Truth proposals} \\ \cline{3-8} 
\multicolumn{1}{c}{}                        & \multicolumn{1}{c}{}                          & \multicolumn{1}{c}{B1}    & \multicolumn{1}{c}{B2}    & \multicolumn{1}{c}{B3}   & \multicolumn{1}{c}{B4}   & \multicolumn{1}{c}{M}    & \multicolumn{1}{c}{C}     \\ \hline
PDVC~\cite{wang2021end}                                         & C3D                                            & 23.09                      & 11.63                      & 5.38                      & 1.95                      & 10.27                     & 19.16                      \\
UEDVC~\cite{zhang2022unifying}                                         & C3D                                         & 28.91                      &  16.54                     & 9.58                     &  5.49                    &10.59                       & 14.64                      \\
PDVC~\cite{wang2021end}                                        & I3D                                            & 23.64                      & 12.58                      & 5.90                      & 2.52                      & 10.56                     & 22.65                      \\ 
\multicolumn{1}{c}{UEDVC}~\cite{zhang2022unifying}                  & \multicolumn{1}{c}{I3D}                       & \multicolumn{1}{c}{26.00} & \multicolumn{1}{c}{15.47} & \multicolumn{1}{c}{9.53} & \multicolumn{1}{c}{5.71} & \multicolumn{1}{c}{10.62} & \multicolumn{1}{c}{18.83} \\ \hline
\end{tabular}
  }
\end{table}

\textbf{Implementation Settings.} For the three models used here, we utilize pre-trained C3D~\cite{tran2015learning} and I3D~\cite{carreira2017quo} models for video feature extraction. In the TDA-CG model, the maximum frame length is set to 300, and any frames beyond this length will be truncated. For the PDVC model, the number of event queries is 100.  For the UEDVC model, the maximum frame length is set to 200. 

\textbf{Results and Analysis.} The localization performance of the model on the UCA dataset is illustrated in Table~\ref{dense event}. Overall, the model exhibits a tendency for relatively higher recall and relatively lower precision. Across different IoU thresholds, the PDVC method consistently outperforms the TDA-CG method, implying the PDVC method's notable competence in target localization. Dense captioning performance is shown in Table~\ref{dense predicted} and Table~\ref{dense ground}. Using predicted proposals, PDVC surpasses TDA-CG across most metrics, especially in BLEU and METEOR scores, implying superior n-gram matching and semantic alignment in PDVC-generated captions. However, all methods exhibit suboptimal CIDEr performance, suggesting room for enhancing caption diversity. In terms of the SODA\_c metric, PDVC's captions better capture the overall video story to some extent compared to TDA-CG. With ground-truth proposals, PDVC outperforms UEDVC in the CIDEr indicator, suggesting a better similarity between its predictions and the ground truth. DVC tasks in UCA encounter challenges from both TSGV and VC, including learning comprehensive temporal information, retaining crucial video event features, and exploring text information. \\

\vspace{-1em}
\subsection{Multimodal Anomaly Detection}
\begin{figure}
    \centering
    \includegraphics[width=.9\linewidth]{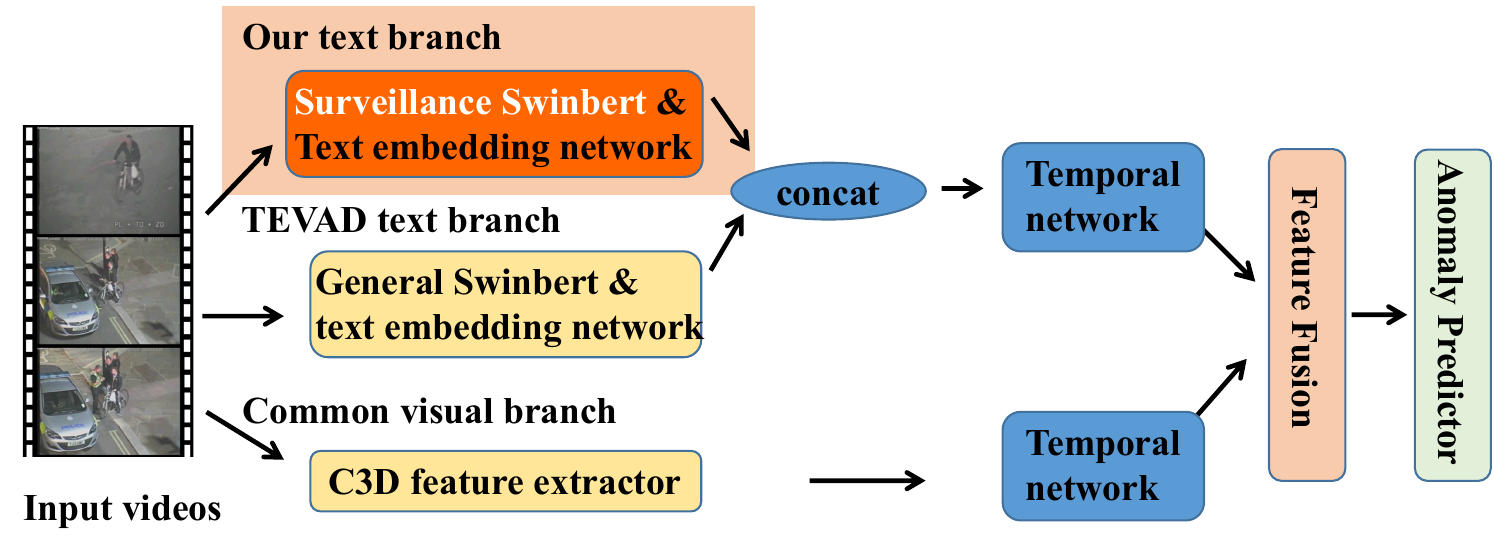}
    \caption{Our enhanced TEVAD framework for multimodal anomaly detection.}
    \vspace{-1em}
    \label{fig:framework}
\end{figure}
\textbf{Task.}
Anomaly detection is one of the important learning tasks in video surveillance. Previous studies ~\cite{liu2019exploring} ~\cite{liu2020enhancing}, have predominantly focused on spatial and temporal features. However, in many complex real-world surveillance videos, these visual features are insufficient in capturing rich semantic meanings. Therefore, Chen \kETAL~\cite{chen2023tevad} proposed the learning task of multimodal anomaly detection~(MAD), which utilizes captions generated by a video captioning model( \ie, Swinbert~\cite{lin2022swinbert}) as a text feature source to improve the performance of traditional anomaly detection.

\textbf{Metric.}
As in TEVAD~\cite{chen2023tevad}, we also employ the micro-averaged AUC (Area Under the ROC curve) by concatenating all frames of the videos to obtain the AUC scores.

\textbf{Baselines and Ours.}
We choose TEVAD~\cite{chen2023tevad} as the baselines of MAD, which fused multi-modal features, including visual features, temporal features, and caption features. We also propose a new framework for anomaly detection with captions on the basis of TEVAD, which is denoted as enhanced TEVAD, and the framework is shown in Figure ~\ref{fig:framework}. TEVAD~\cite{chen2023tevad} deploys Swinbert~\cite{lin2022swinbert} pre-trained on VATEX~\cite{Wang2019VaTeXAL} (General Swinbert in Figure~\ref{fig:framework}) as the video caption extractors. VATEX is a large-scale and general video dataset, which provides a general video captioning capacity. Differently, our pre-trained Swinbert on UCA can provide more capacity for captioning the surveillance videos. On the basis of TEVAD, our enhanced TEVAD further introduces our pre-trained Swinbert on UCA (Surveillance Swinbert in Figure~\ref{fig:framework}, as a plug-and-play module) as a domain-specific feature branch for anomaly detection.

\textbf{Implementation Settings.} We use the original experiment settings in TEVAD, and deploy feature concat to fuse multimodal features. The main difference between TEVAD and our enhanced TEVAD is that we introduce a new Surveillance Swinbert trained on UCA. 

\begin{figure}[h]
    \centering
    \includegraphics[width=0.9\linewidth]{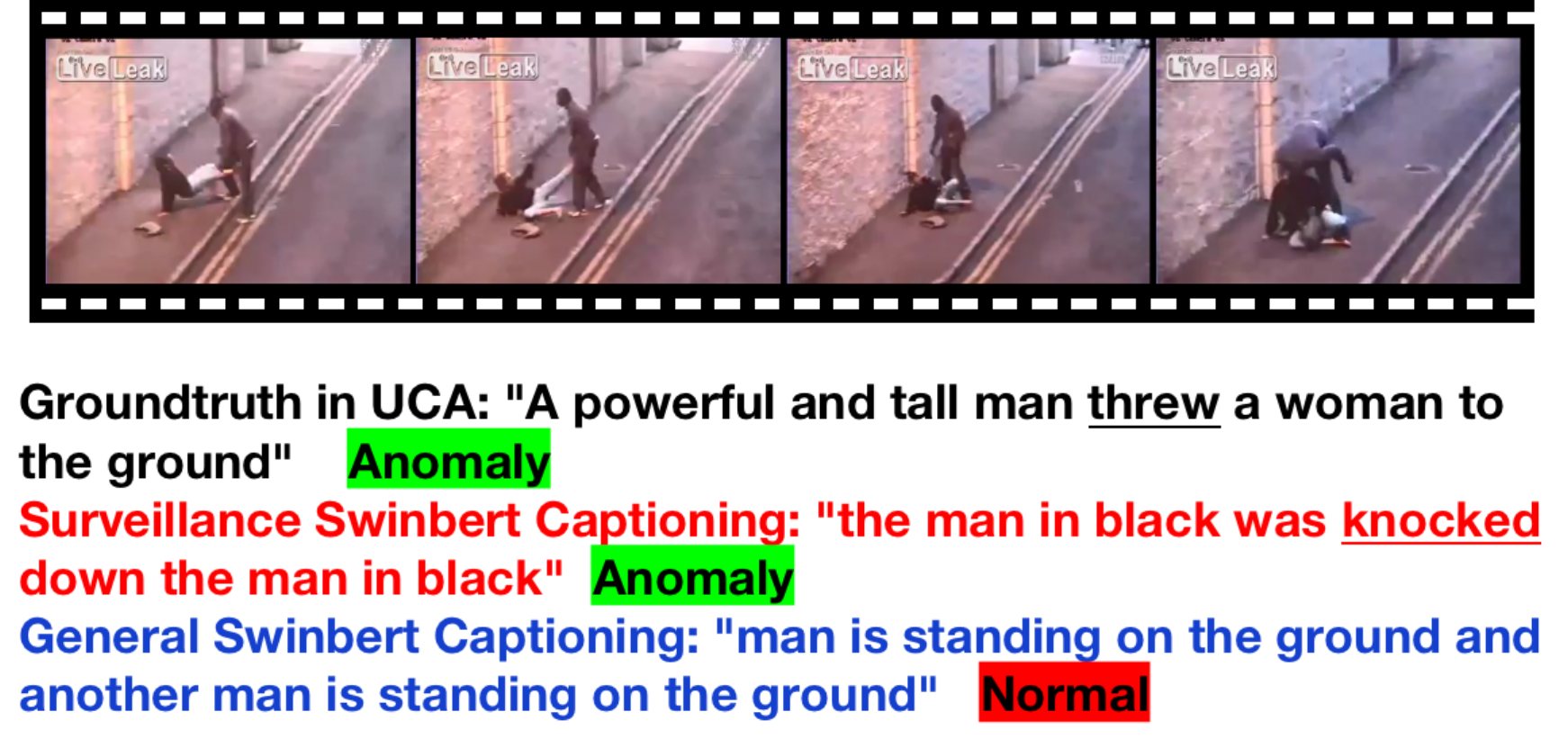}
    \vspace{-1em}
    \caption{Examples of different video captioning results in MAD.}
    \label{fig:MAD}
\end{figure}
\vspace{-1em}
\textbf{Results and Analysis.}
The results in Table~\ref{ADWC} show our enhanced TEVAD outperforms others. It indicates that our pre-trained Swinbert on UCA can improve anomaly detection accuracy by providing surveillance-domain-specific feature information. In the example of Figure~\ref{fig:MAD}, we can find that our Surveillance Swinbert can generate captions with anomalous descriptions, which is useful in multimodal anomaly detection. This also demonstrates the effectiveness of our UCA and our trained multimodal learning models, which can also play an important role in refining existing surveillance tasks.

\begin{table}[h]
\centering
\vspace{-0.5em}
\caption{Multimodal Anomaly detection (MAD) on UCF-Crime, where "visual" denotes visual features.}
\vspace{-0.5em}
\label{ADWC}
\resizebox{1.\linewidth}{!}{
\begin{tabular}{ccccc}
\toprule
method & visual & General Swinbert & Surveillance Swinbert & AUC \\
\midrule
Visual &\checkmark & \ding{55}& \ding{55} & 83.1\% \\
TEVAD~\cite{chen2023tevad} (2023)& \checkmark & \checkmark &\ding{55} & 84.9\% \\
Enhanced TEVAD (Ours)& \checkmark & \checkmark& \checkmark & 85.3\%
 \\ \bottomrule
\end{tabular}
}
\end{table}


\subsection{Discussion}

\begin{figure}[ht]
    \centering
    \includegraphics[width=1.\linewidth]{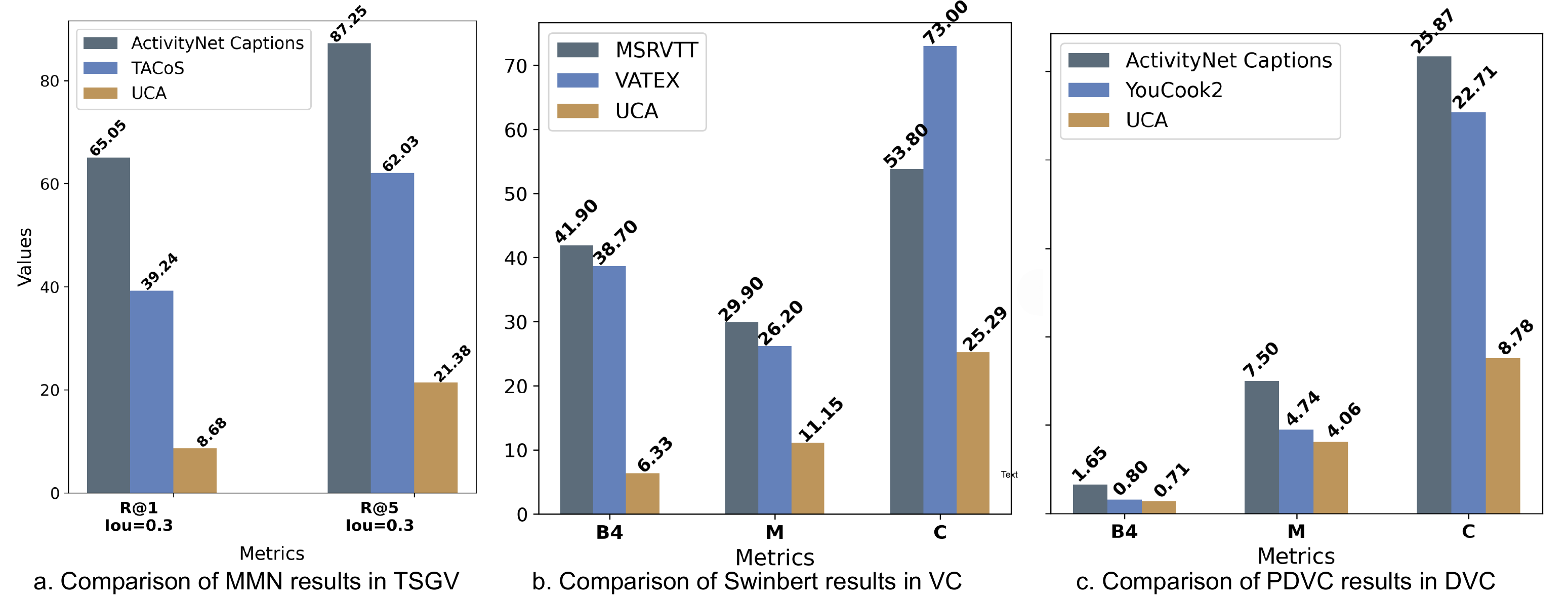}

    \caption{Illustrations of the performance gaps lie in our UCA and other conventional datasets.}
    \vspace{-1em}
    \label{fig:data_com}
\end{figure}
We have conducted experiments on mainstream multimodal video understanding tasks. It is the first time that these tasks are oriented to surveillance video. Through these experiments, we can provide a benchmark for surveillance video-and-language
understanding. By analyzing the experiment results, we give the following discussions.

\begin{itemize}
    \item  SOTA methods oriented to conventional videos cannot perform well on our new UCA, which demonstrates the challenges of surveillance video-and-language understanding. Several instances of these performance gaps lie in multimodal learning between UCA and conventional video datasets are shown in Figure~\ref{fig:data_com}. For TSGV tasks, surveillance videos have long video lengths, leading to the difficulties of temporal grounding. For VC and DVC tasks, due to different domains of videos having particularities, a general model cannot well adapt to a surveillance domain.
    \item  We also recognize that the basic learning model architecture must be modified and designed based on the unique characteristics of surveillance video datasets. We suggest that researchers pay more attention to specific aspects such as video-and-language modality alignment, video-temporal information feature fusion, and more comprehensive semantic information mining in the field of surveillance videos. These areas hold significant potential for improving the understanding of surveillance video content.
    \item For the multimodal anomaly detection task, we can find that surveillance video captioning can play an auxiliary role in anomaly detection by fusing the caption features. The model trained on our UCA can complement the specific capacity of surveillance video understanding, compared with the model trained on conventional videos. This highlights the effectiveness of our UCA. Furthermore, a well-trained multimodal surveillance video learning model can become a promising factor in lifting existing surveillance and security tasks. In the future, we will pay more attention to mining more potentials of our annotations. 
\end{itemize}
However, obtaining such a well-trained surveillance learning model needs the support of datasets. Because the collection of surveillance video data is still challenging, our UCA based on UCF-Crime, also suffers from insufficient data volumes. In the future, we can make more annotations on the newly released surveillance videos, \eg, NWPU~\cite{Cao_2023_CVPR}.

\section{Conclusions}
In the current AI field, there is a significant gap in research on multimodal surveillance video datasets, despite their potential for contributing to social security and daily life. To bridge this gap, we propose UCA, as the first multimodal surveillance video dataset, which is derived from re-annotating UCF-Crime. Our annotation focuses on providing detailed event descriptions and start-stop time marks for normal or abnormal events occurring in surveillance videos, resulting in the creation of a multimodal surveillance video dataset with 23,542 sentence-level descriptions and frame-level time records. Moreover, we conducted experiments on 4 multimodal tasks using the UCA dataset, evaluating the performance of 17 benchmark methods. Based on the evaluations, we identify significant opportunities and challenges in this field and emphasize the need for further research in both the dataset and model and aspects.

{
    \small
    \bibliographystyle{ieeenat_fullname}
    \bibliography{main}

\begin{thebibliography}{54}
\providecommand{\natexlab}[1]{#1}
\providecommand{\url}[1]{\texttt{#1}}
\expandafter\ifx\csname urlstyle\endcsname\relax
  \providecommand{\doi}[1]{doi: #1}\else
  \providecommand{\doi}{doi: \begingroup \urlstyle{rm}\Url}\fi

\bibitem[Abdar et~al.(2023)Abdar, Kollati, Kuraparthi, Pourpanah, McDuff, Ghavamzadeh, Yan, Mohamed, Khosravi, Cambria, et~al.]{abdar2023review}
Moloud Abdar, Meenakshi Kollati, Swaraja Kuraparthi, Farhad Pourpanah, Daniel McDuff, Mohammad Ghavamzadeh, Shuicheng Yan, Abduallah Mohamed, Abbas Khosravi, Erik Cambria, et~al.
\newblock A review of deep learning for video captioning.
\newblock \emph{arXiv preprint arXiv:2304.11431}, 2023.

\bibitem[Adam et~al.(2008)Adam, Rivlin, Shimshoni, and Reinitz]{adam2008robust}
Amit Adam, Ehud Rivlin, Ilan Shimshoni, and Daviv Reinitz.
\newblock Robust real-time unusual event detection using multiple fixed-location monitors.
\newblock \emph{IEEE transactions on pattern analysis and machine intelligence}, 30\penalty0 (3):\penalty0 555--560, 2008.

\bibitem[Anne~Hendricks et~al.(2017)Anne~Hendricks, Wang, Shechtman, Sivic, Darrell, and Russell]{anne2017localizing}
Lisa Anne~Hendricks, Oliver Wang, Eli Shechtman, Josef Sivic, Trevor Darrell, and Bryan Russell.
\newblock Localizing moments in video with natural language.
\newblock In \emph{Proceedings of the IEEE international conference on computer vision}, pages 5803--5812, 2017.

\bibitem[Cao et~al.(2023)Cao, Lu, Wang, and Zhang]{Cao_2023_CVPR}
Congqi Cao, Yue Lu, Peng Wang, and Yanning Zhang.
\newblock A new comprehensive benchmark for semi-supervised video anomaly detection and anticipation.
\newblock In \emph{Proceedings of the IEEE/CVF Conference on Computer Vision and Pattern Recognition (CVPR)}, pages 20392--20401, 2023.

\bibitem[Carreira and Zisserman(2017)]{carreira2017quo}
Joao Carreira and Andrew Zisserman.
\newblock Quo vadis, action recognition? a new model and the kinetics dataset.
\newblock In \emph{proceedings of the IEEE Conference on Computer Vision and Pattern Recognition}, pages 6299--6308, 2017.

\bibitem[Chen and Dolan(2011)]{chen2011collecting}
David Chen and William~B Dolan.
\newblock Collecting highly parallel data for paraphrase evaluation.
\newblock In \emph{Proceedings of the 49th annual meeting of the association for computational linguistics: human language technologies}, pages 190--200, 2011.

\bibitem[Chen et~al.(2019)Chen, Yao, and Jiang]{chen2019deep}
Shaoxiang Chen, Ting Yao, and Yu-Gang Jiang.
\newblock Deep learning for video captioning: A review.
\newblock In \emph{IJCAI}, page~2, 2019.

\bibitem[Chen et~al.(2023{\natexlab{a}})Chen, Ma, Yew, Hur, and Khoo]{chen2023tevad}
Weiling Chen, Keng~Teck Ma, Zi~Jian Yew, Minhoe Hur, and David Aik-Aun Khoo.
\newblock Tevad: Improved video anomaly detection with captions.
\newblock In \emph{Proceedings of the IEEE/CVF Conference on Computer Vision and Pattern Recognition}, pages 5548--5558, 2023{\natexlab{a}}.

\bibitem[Chen et~al.(2023{\natexlab{b}})Chen, Liu, Zhang, Fok, Qi, and Wu]{chen2023mgfn}
Yingxian Chen, Zhengzhe Liu, Baoheng Zhang, Wilton Fok, Xiaojuan Qi, and Yik-Chung Wu.
\newblock Mgfn: Magnitude-contrastive glance-and-focus network for weakly-supervised video anomaly detection.
\newblock In \emph{Proceedings of the AAAI Conference on Artificial Intelligence}, pages 387--395, 2023{\natexlab{b}}.

\bibitem[Das et~al.(2013)Das, Xu, Doell, and Corso]{das2013thousand}
Pradipto Das, Chenliang Xu, Richard~F Doell, and Jason~J Corso.
\newblock A thousand frames in just a few words: Lingual description of videos through latent topics and sparse object stitching.
\newblock In \emph{Proceedings of the IEEE conference on computer vision and pattern recognition}, pages 2634--2641, 2013.

\bibitem[Denkowski and Lavie(2014)]{denkowski2014meteor}
Michael Denkowski and Alon Lavie.
\newblock Meteor universal: Language specific translation evaluation for any target language.
\newblock In \emph{Proceedings of the ninth workshop on statistical machine translation}, pages 376--380, 2014.

\bibitem[Gao et~al.(2017)Gao, Sun, Yang, and Nevatia]{gao2017tall}
Jiyang Gao, Chen Sun, Zhenheng Yang, and Ram Nevatia.
\newblock Tall: Temporal activity localization via language query.
\newblock In \emph{Proceedings of the IEEE international conference on computer vision}, pages 5267--5275, 2017.

\bibitem[Hara et~al.(2018)Hara, Kataoka, and Satoh]{hara2018can}
Kensho Hara, Hirokatsu Kataoka, and Yutaka Satoh.
\newblock Can spatiotemporal 3d cnns retrace the history of 2d cnns and imagenet?
\newblock In \emph{Proceedings of the IEEE conference on Computer Vision and Pattern Recognition}, pages 6546--6555, 2018.

\bibitem[He et~al.(2019)He, Zhao, Huang, Li, Liu, and Wen]{he2019read}
Dongliang He, Xiang Zhao, Jizhou Huang, Fu Li, Xiao Liu, and Shilei Wen.
\newblock Read, watch, and move: Reinforcement learning for temporally grounding natural language descriptions in videos.
\newblock In \emph{Proceedings of the AAAI Conference on Artificial Intelligence}, pages 8393--8400, 2019.

\bibitem[Kiros et~al.(2015)Kiros, Zhu, Salakhutdinov, Zemel, Urtasun, Torralba, and Fidler]{kiros2015skip}
Ryan Kiros, Yukun Zhu, Russ~R Salakhutdinov, Richard Zemel, Raquel Urtasun, Antonio Torralba, and Sanja Fidler.
\newblock Skip-thought vectors.
\newblock \emph{Advances in neural information processing systems}, 28, 2015.

\bibitem[Krishna et~al.(2017)Krishna, Hata, Ren, Fei-Fei, and Carlos~Niebles]{krishna2017dense}
Ranjay Krishna, Kenji Hata, Frederic Ren, Li Fei-Fei, and Juan Carlos~Niebles.
\newblock Dense-captioning events in videos.
\newblock In \emph{Proceedings of the IEEE international conference on computer vision}, pages 706--715, 2017.

\bibitem[Lan et~al.(2023)Lan, Yuan, Wang, Wang, and Zhu]{lan2023survey}
Xiaohan Lan, Yitian Yuan, Xin Wang, Zhi Wang, and Wenwu Zhu.
\newblock A survey on temporal sentence grounding in videos.
\newblock \emph{ACM Transactions on Multimedia Computing, Communications and Applications}, 19\penalty0 (2):\penalty0 1--33, 2023.

\bibitem[Li et~al.(2023)Li, Xie, Xie, Zhao, Zhang, Zheng, Zhao, and Zhang]{li2023momentdiff}
Pandeng Li, Chen-Wei Xie, Hongtao Xie, Liming Zhao, Lei Zhang, Yun Zheng, Deli Zhao, and Yongdong Zhang.
\newblock Momentdiff: Generative video moment retrieval from random to real.
\newblock \emph{arXiv preprint arXiv:2307.02869}, 2023.

\bibitem[Li et~al.(2019)Li, Tao, Li, and Fu]{li2019visual}
Sheng Li, Zhiqiang Tao, Kang Li, and Yun Fu.
\newblock Visual to text: Survey of image and video captioning.
\newblock \emph{IEEE Transactions on Emerging Topics in Computational Intelligence}, 3\penalty0 (4):\penalty0 297--312, 2019.

\bibitem[Li et~al.(2013)Li, Mahadevan, and Vasconcelos]{li2013anomaly}
Weixin Li, Vijay Mahadevan, and Nuno Vasconcelos.
\newblock Anomaly detection and localization in crowded scenes.
\newblock \emph{IEEE transactions on pattern analysis and machine intelligence}, 36\penalty0 (1):\penalty0 18--32, 2013.

\bibitem[Lin(2004)]{lin2004rouge}
Chin-Yew Lin.
\newblock Rouge: A package for automatic evaluation of summaries.
\newblock In \emph{Text summarization branches out}, pages 74--81, 2004.

\bibitem[Lin et~al.(2022)Lin, Li, Lin, Ahmed, Gan, Liu, Lu, and Wang]{lin2022swinbert}
Kevin Lin, Linjie Li, Chung-Ching Lin, Faisal Ahmed, Zhe Gan, Zicheng Liu, Yumao Lu, and Lijuan Wang.
\newblock Swinbert: End-to-end transformers with sparse attention for video captioning.
\newblock In \emph{Proceedings of the IEEE/CVF Conference on Computer Vision and Pattern Recognition}, pages 17949--17958, 2022.

\bibitem[Liu and Ma(2019)]{liu2019exploring}
Kun Liu and Huadong Ma.
\newblock Exploring background-bias for anomaly detection in surveillance videos.
\newblock In \emph{Proceedings of the 27th ACM International Conference on Multimedia}, pages 1490--1499, 2019.

\bibitem[Liu et~al.(2020)Liu, Zhu, Fu, Ma, and Chua]{liu2020enhancing}
Kun Liu, Minzhi Zhu, Huiyuan Fu, Huadong Ma, and Tat-Seng Chua.
\newblock Enhancing anomaly detection in surveillance videos with transfer learning from action recognition.
\newblock In \emph{Proceedings of the 28th ACM International Conference on Multimedia}, pages 4664--4668, 2020.

\bibitem[Liu et~al.(2023)Liu, Nie, Wang, Wang, and Rui]{liu2023survey}
Meng Liu, Liqiang Nie, Yunxiao Wang, Meng Wang, and Yong Rui.
\newblock A survey on video moment localization.
\newblock \emph{ACM Computing Surveys}, 55\penalty0 (9):\penalty0 1--37, 2023.

\bibitem[Liu et~al.(2022)Liu, Ning, Cao, Wei, Zhang, Lin, and Hu]{liu2022video}
Ze Liu, Jia Ning, Yue Cao, Yixuan Wei, Zheng Zhang, Stephen Lin, and Han Hu.
\newblock Video swin transformer.
\newblock In \emph{Proceedings of the IEEE/CVF conference on computer vision and pattern recognition}, pages 3202--3211, 2022.

\bibitem[Lu et~al.(2013)Lu, Shi, and Jia]{lu2013abnormal}
Cewu Lu, Jianping Shi, and Jiaya Jia.
\newblock Abnormal event detection at 150 fps in matlab.
\newblock In \emph{Proceedings of the IEEE international conference on computer vision}, pages 2720--2727, 2013.

\bibitem[Luo et~al.(2017)Luo, Liu, and Gao]{Luo_2017_ICCV}
Weixin Luo, Wen Liu, and Shenghua Gao.
\newblock A revisit of sparse coding based anomaly detection in stacked rnn framework.
\newblock In \emph{Proceedings of the IEEE International Conference on Computer Vision (ICCV)}, 2017.

\bibitem[Mittal et~al.(2022)Mittal, Ghosal, and Bansal]{mittal2022savchoi}
Ansh Mittal, Shuvam Ghosal, and Rishibha Bansal.
\newblock Savchoi: Detecting suspicious activities using dense video captioning with human object interactions.
\newblock \emph{arXiv preprint arXiv:2207.11838}, 2022.

\bibitem[Mun et~al.(2020)Mun, Cho, and Han]{mun2020local}
Jonghwan Mun, Minsu Cho, and Bohyung Han.
\newblock Local-global video-text interactions for temporal grounding.
\newblock In \emph{Proceedings of the IEEE/CVF Conference on Computer Vision and Pattern Recognition}, pages 10810--10819, 2020.

\bibitem[Papineni et~al.(2002)Papineni, Roukos, Ward, and Zhu]{papineni2002bleu}
Kishore Papineni, Salim Roukos, Todd Ward, and Wei-Jing Zhu.
\newblock Bleu: a method for automatic evaluation of machine translation.
\newblock In \emph{Proceedings of the 40th annual meeting of the Association for Computational Linguistics}, pages 311--318, 2002.

\bibitem[Pei et~al.(2019)Pei, Zhang, Wang, Ke, Shen, and Tai]{pei2019memory}
Wenjie Pei, Jiyuan Zhang, Xiangrong Wang, Lei Ke, Xiaoyong Shen, and Yu-Wing Tai.
\newblock Memory-attended recurrent network for video captioning.
\newblock In \emph{Proceedings of the IEEE/CVF Conference on Computer Vision and Pattern Recognition}, pages 8347--8356, 2019.

\bibitem[Regneri et~al.(2013)Regneri, Rohrbach, Wetzel, Thater, Schiele, and Pinkal]{regneri2013grounding}
Michaela Regneri, Marcus Rohrbach, Dominikus Wetzel, Stefan Thater, Bernt Schiele, and Manfred Pinkal.
\newblock Grounding action descriptions in videos.
\newblock \emph{Transactions of the Association for Computational Linguistics}, 1:\penalty0 25--36, 2013.

\bibitem[Rohrbach et~al.(2015)Rohrbach, Rohrbach, Tandon, and Schiele]{rohrbach2015dataset}
Anna Rohrbach, Marcus Rohrbach, Niket Tandon, and Bernt Schiele.
\newblock A dataset for movie description.
\newblock In \emph{Proceedings of the IEEE conference on computer vision and pattern recognition}, pages 3202--3212, 2015.

\bibitem[Ryu et~al.(2021)Ryu, Kang, Kang, and Yoo]{ryu2021semantic}
Hobin Ryu, Sunghun Kang, Haeyong Kang, and Chang~D Yoo.
\newblock Semantic grouping network for video captioning.
\newblock In \emph{proceedings of the AAAI Conference on Artificial Intelligence}, pages 2514--2522, 2021.

\bibitem[{\c{S}}eng{\"o}n{\"u}l et~al.(2023){\c{S}}eng{\"o}n{\"u}l, Samet, Abu Al-Haija, Alqahtani, Alturki, and Alsulami]{csengonul2023analysis}
Erkan {\c{S}}eng{\"o}n{\"u}l, Refik Samet, Qasem Abu Al-Haija, Ali Alqahtani, Badraddin Alturki, and Abdulaziz~A Alsulami.
\newblock An analysis of artificial intelligence techniques in surveillance video anomaly detection: A comprehensive survey.
\newblock \emph{Applied Sciences}, 13\penalty0 (8):\penalty0 4956, 2023.

\bibitem[Shen et~al.(2023)Shen, Gu, Xu, Fan, Wen, and Zhang]{shen2023accurate}
Yaojie Shen, Xin Gu, Kai Xu, Heng Fan, Longyin Wen, and Libo Zhang.
\newblock Accurate and fast compressed video captioning.
\newblock In \emph{Proceedings of the IEEE/CVF International Conference on Computer Vision}, pages 15558--15567, 2023.

\bibitem[Sigurdsson et~al.(2016)Sigurdsson, Varol, Wang, Farhadi, Laptev, and Gupta]{sigurdsson2016hollywood}
Gunnar~A Sigurdsson, G{\"u}l Varol, Xiaolong Wang, Ali Farhadi, Ivan Laptev, and Abhinav Gupta.
\newblock Hollywood in homes: Crowdsourcing data collection for activity understanding.
\newblock In \emph{Computer Vision--ECCV 2016: 14th European Conference, Amsterdam, The Netherlands, October 11--14, 2016, Proceedings, Part I 14}, pages 510--526. Springer, 2016.

\bibitem[Sultani et~al.(2018)Sultani, Chen, and Shah]{sultani2018real}
Waqas Sultani, Chen Chen, and Mubarak Shah.
\newblock Real-world anomaly detection in surveillance videos.
\newblock In \emph{Proceedings of the IEEE conference on computer vision and pattern recognition}, pages 6479--6488, 2018.

\bibitem[Tran et~al.(2015)Tran, Bourdev, Fergus, Torresani, and Paluri]{tran2015learning}
Du Tran, Lubomir Bourdev, Rob Fergus, Lorenzo Torresani, and Manohar Paluri.
\newblock Learning spatiotemporal features with 3d convolutional networks.
\newblock In \emph{Proceedings of the IEEE international conference on computer vision}, pages 4489--4497, 2015.

\bibitem[Vedantam et~al.(2015)Vedantam, Lawrence~Zitnick, and Parikh]{vedantam2015cider}
Ramakrishna Vedantam, C Lawrence~Zitnick, and Devi Parikh.
\newblock Cider: Consensus-based image description evaluation.
\newblock In \emph{Proceedings of the IEEE conference on computer vision and pattern recognition}, pages 4566--4575, 2015.

\bibitem[Venugopalan et~al.(2015)Venugopalan, Rohrbach, Donahue, Mooney, Darrell, and Saenko]{venugopalan2015sequence}
Subhashini Venugopalan, Marcus Rohrbach, Jeffrey Donahue, Raymond Mooney, Trevor Darrell, and Kate Saenko.
\newblock Sequence to sequence-video to text.
\newblock In \emph{Proceedings of the IEEE international conference on computer vision}, pages 4534--4542, 2015.

\bibitem[Wang et~al.(2018{\natexlab{a}})Wang, Ma, Zhang, and Liu]{wang2018reconstruction}
Bairui Wang, Lin Ma, Wei Zhang, and Wei Liu.
\newblock Reconstruction network for video captioning.
\newblock In \emph{Proceedings of the IEEE conference on computer vision and pattern recognition}, pages 7622--7631, 2018{\natexlab{a}}.

\bibitem[Wang et~al.(2018{\natexlab{b}})Wang, Jiang, Ma, Liu, and Xu]{wang2018bidirectional}
Jingwen Wang, Wenhao Jiang, Lin Ma, Wei Liu, and Yong Xu.
\newblock Bidirectional attentive fusion with context gating for dense video captioning.
\newblock In \emph{CVPR}, 2018{\natexlab{b}}.

\bibitem[Wang et~al.(2021)Wang, Zhang, Lu, Zheng, Cheng, and Luo]{wang2021end}
Teng Wang, Ruimao Zhang, Zhichao Lu, Feng Zheng, Ran Cheng, and Ping Luo.
\newblock End-to-end dense video captioning with parallel decoding.
\newblock In \emph{Proceedings of the IEEE/CVF International Conference on Computer Vision}, pages 6847--6857, 2021.

\bibitem[Wang et~al.(2019)Wang, Wu, Chen, Lei, Wang, and Wang]{Wang2019VaTeXAL}
Xin Wang, Jiawei Wu, Junkun Chen, Li Lei, Yuan-Fang Wang, and William~Yang Wang.
\newblock Vatex: A large-scale, high-quality multilingual dataset for video-and-language research.
\newblock In \emph{Proceedings of the IEEE/CVF International Conference on Computer Vision}, pages 4580--4590, 2019.

\bibitem[Wang et~al.(2022)Wang, Wang, Wu, Li, and Wu]{wang2022negative}
Zhenzhi Wang, Limin Wang, Tao Wu, Tianhao Li, and Gangshan Wu.
\newblock Negative sample matters: A renaissance of metric learning for temporal grounding.
\newblock In \emph{Proceedings of the AAAI Conference on Artificial Intelligence}, pages 2613--2623, 2022.

\bibitem[Xu et~al.(2016)Xu, Mei, Yao, and Rui]{xu2016msr}
Jun Xu, Tao Mei, Ting Yao, and Yong Rui.
\newblock Msr-vtt: A large video description dataset for bridging video and language.
\newblock In \emph{Proceedings of the IEEE conference on computer vision and pattern recognition}, pages 5288--5296, 2016.

\bibitem[Yuan et~al.(2019)Yuan, Ma, Wang, Liu, and Zhu]{yuan2019semantic}
Yitian Yuan, Lin Ma, Jingwen Wang, Wei Liu, and Wenwu Zhu.
\newblock Semantic conditioned dynamic modulation for temporal sentence grounding in videos.
\newblock \emph{Advances in Neural Information Processing Systems}, 32, 2019.

\bibitem[Zhang et~al.(2022{\natexlab{a}})Zhang, Sun, Jing, and Zhou]{zhang2022temporal}
Hao Zhang, Aixin Sun, Wei Jing, and Joey~Tianyi Zhou.
\newblock Temporal sentence grounding in videos: A survey and future directions.
\newblock \emph{arXiv preprint arXiv:2201.08071}, 2022{\natexlab{a}}.

\bibitem[Zhang et~al.(2022{\natexlab{b}})Zhang, Song, and Jin]{zhang2022unifying}
Qi Zhang, Yuqing Song, and Qin Jin.
\newblock Unifying event detection and captioning as sequence generation via pre-training.
\newblock In \emph{Computer Vision--ECCV 2022: 17th European Conference, Tel Aviv, Israel, October 23--27, 2022, Proceedings, Part XXXVI}, pages 363--379. Springer, 2022{\natexlab{b}}.

\bibitem[Zhang et~al.(2020)Zhang, Peng, Fu, and Luo]{zhang2020learning}
Songyang Zhang, Houwen Peng, Jianlong Fu, and Jiebo Luo.
\newblock Learning 2d temporal adjacent networks for moment localization with natural language.
\newblock In \emph{Proceedings of the AAAI Conference on Artificial Intelligence}, pages 12870--12877, 2020.

\bibitem[Zhang et~al.(2021)Zhang, Peng, Fu, Lu, and Luo]{zhang2021multi}
Songyang Zhang, Houwen Peng, Jianlong Fu, Yijuan Lu, and Jiebo Luo.
\newblock Multi-scale 2d temporal adjacency networks for moment localization with natural language.
\newblock \emph{IEEE Transactions on Pattern Analysis and Machine Intelligence}, 44\penalty0 (12):\penalty0 9073--9087, 2021.

\bibitem[Zhou et~al.(2023)Zhou, Yu, and Yang]{zhou2023dual}
Hang Zhou, Junqing Yu, and Wei Yang.
\newblock Dual memory units with uncertainty regulation for weakly supervised video anomaly detection.
\newblock \emph{arXiv preprint arXiv:2302.05160}, 2023.

\end{thebibliography}
}

\appendix

\clearpage
\setcounter{page}{1}
\setcounter{section}{0}

\section*{Appendix}
The Appendix consists of four parts:
\begin{itemize}
    \item Sec. A: This section presents the details of dataset construction, which includes video collection, annotation details, annotation guidance, and analysis of annotation agreement.
    \item Sec. B: This section provides experiment details for the submitted manuscript, including a thorough explanation of the experiment settings and visualizations for three multimodal learning tasks.
    \item Sec. C: In this section, additional experiments and results are provided, covering the performance on normal and anomalous videos, performance with gender-neutral annotations, and anomaly detection.
    \item Sec. D: This section clarifies the license details and accessibility of our UCA dataset.
\end{itemize}

\section{Dataset Construction Details}
\subsection{Video Collection}

All the videos in our UCA dataset are sourced from UCF-Crime~\cite{sultani2018real}, a real-world surveillance dataset released at CVPR 2018. The dataset encompasses a wide range of event categories, including Abuse, Arrest, Arson, Assault, Burglary, Explosion, Fighting, Road Accidents, Robbery, Shooting, Shoplifting, Stealing, Vandalism, and Normal Videos. 

Out of the original 1900 videos, we excluded 46 low-quality videos, which resulted in a final collection of 1854 videos. The criteria for removing these low-quality videos included severe occlusion, blurry content, excessively fast playback speeds, and the presence of duplicate videos. The presence of such low-quality videos poses challenges in subsequent annotation tasks, as they are difficult to identify through manual inspection. The statistics of UCF-Crime and our UCA in terms of the number of videos in different categories are shown in Table~\ref{Data2}.

\begin{table*}[ht]
\caption{The comparison of UCF-Crime and UCA in the video numbers of different categories.}
\label{Data2}%
\centering
\resizebox{0.9\textwidth}{!}{
\begin{tabular}{ccccccccc}
\toprule
\multirow{2}{*}{Video numbers} &
  \multirow{2}{*}{Abuse} &
  \multirow{2}{*}{Arrest} &
  \multirow{2}{*}{Arson} &
  \multirow{2}{*}{Assault} &
  \multirow{2}{*}{Burglary} &
  \multirow{2}{*}{Explosion} &
  \multirow{2}{*}{Fighting} &
  \multirow{2}{*}{Road Accidents} \\
                        &     &    &    &     &     &     &     &      \\ \midrule
UCF-Crime & 50  & 50 & 50 & 50  & 100 & 50  & 50  & 150  \\ 
Our UCA       & 50  & 50 & 50 & 48  & 100 & 50  & 50  & 148  \\ \midrule
Video numbers &
  Robbery &
  Shooting &
  Shoplifting &
  Stealing &
  Vandalism &
  \begin{tabular}[c]{@{}c@{}}Training\_\\ Normal\_Videos\end{tabular} &
  \begin{tabular}[c]{@{}c@{}}Testing\_\\ Normal\_Videos\end{tabular} &
  Summary \\ \midrule
UCF-Crime & 150 & 50 & 50 & 100 & 50  & 800 & 150 & 1900 \\ 
Our UCA       & 149 & 50 & 50 & 100 & 49  & 764 & 146 & 1854 \\ \bottomrule
\end{tabular}%
}
\end{table*}

Moreover, following the partitioning approach of the original UCF-Crime dataset, UCA is categorized into two main groups: "Abnormal" and "Normal" videos, as shown in Table~\ref{tab:abnormal}. In this context, "Abnormal" videos denote those containing scenes with exceptional occurrences or criminal activities present within the original videos. 

\begin{table}[h]
\centering
\caption{Abnormal and normal video splits of our UCA dataset.}
\label{tab:abnormal}
\resizebox{0.97\linewidth}{!}{%
\begin{tabular}{lcccccc}
\toprule
\multirow{2}{*}{Statistics}  & \multicolumn{2}{c}{Train}      & \multicolumn{2}{c}{Val}      & \multicolumn{2}{c}{Test}      \\ \cmidrule{2-7}
              & Abnormal & Normal   &Abnormal & Normal & Abnormal & Normal    \\ \midrule
\#Video     & 576    & 589   & 162 & 217 & 206 & 104                        \\ 
Video length(h)      &  20.94    & 54.59    & 5.88& 15.29 & 6.85 & 18.38    \\ 
 \bottomrule

\end{tabular}}%
\end{table}

\subsection{Annotations}
For videos in UCF-Crime, we provide fine-grained annotations that describe the events occurring within the video. These annotations are highly beneficial for tasks such as video understanding, video temporal localization, and video caption generation. Each annotation also includes the precise start and end times of the events, accurate up to 0.1 seconds. 

To meet the needs of different researchers and enhance the convenience of data processing, we have uploaded two versions of the annotation files in our dataset project at \url{https://xuange923.github.io/Surveillance-Video-Understanding}, namely `txt' and `json' formats. The original annotation data was collected in txt format, which is simple and easy to read, facilitating preliminary data collection and quick reference. Additionally, considering the need for research and experimentation, we have converted these txt annotations into json format for ease of use in subsequent experiments.

\begin{figure}[ht]
    \centering
    \includegraphics[width=\linewidth]{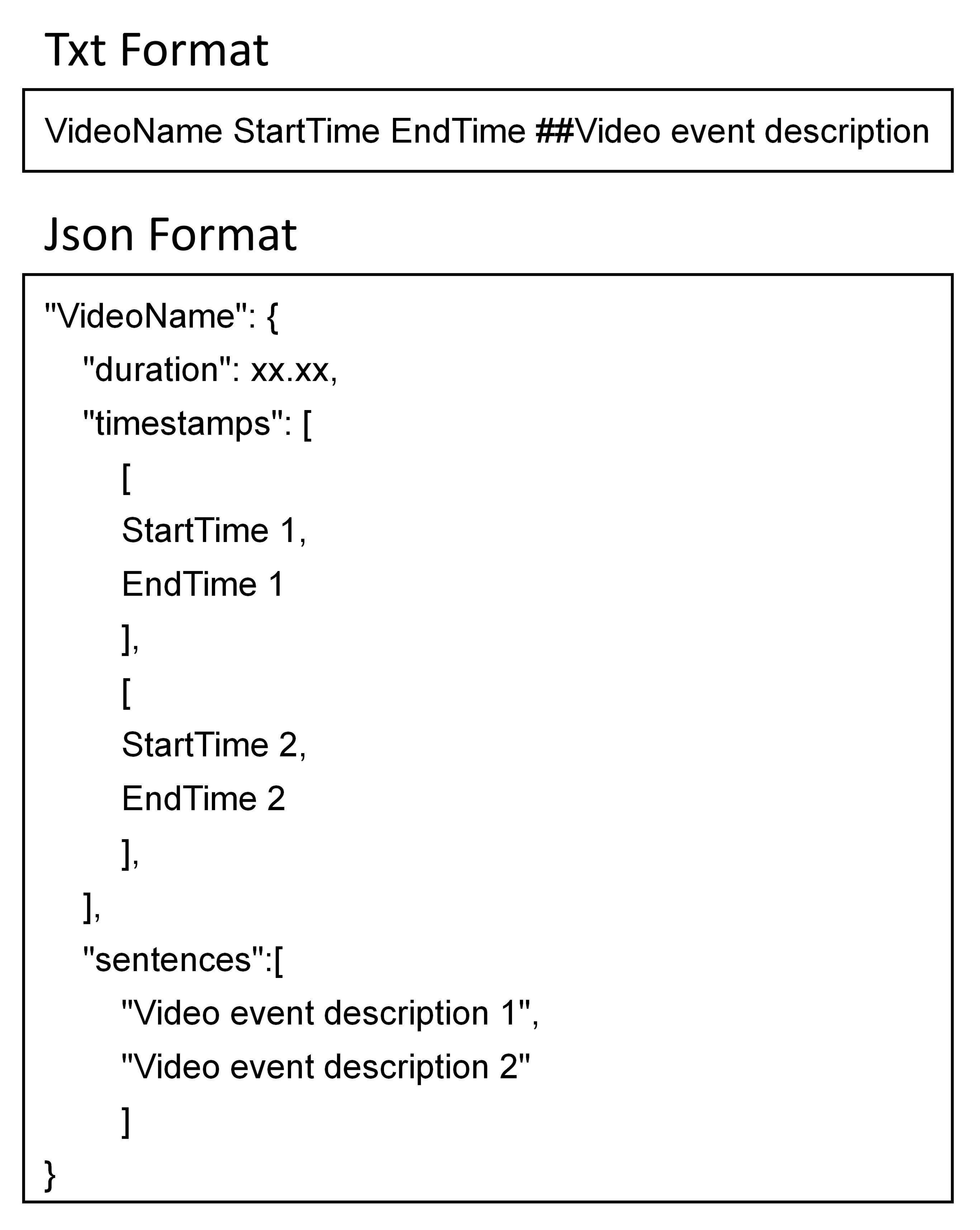}
    \caption{Comparison of Two Versions of Annotation Formats}
    \label{fig-format}
\end{figure}

\begin{figure*}[h]
    \centering
    \includegraphics[width=\linewidth]{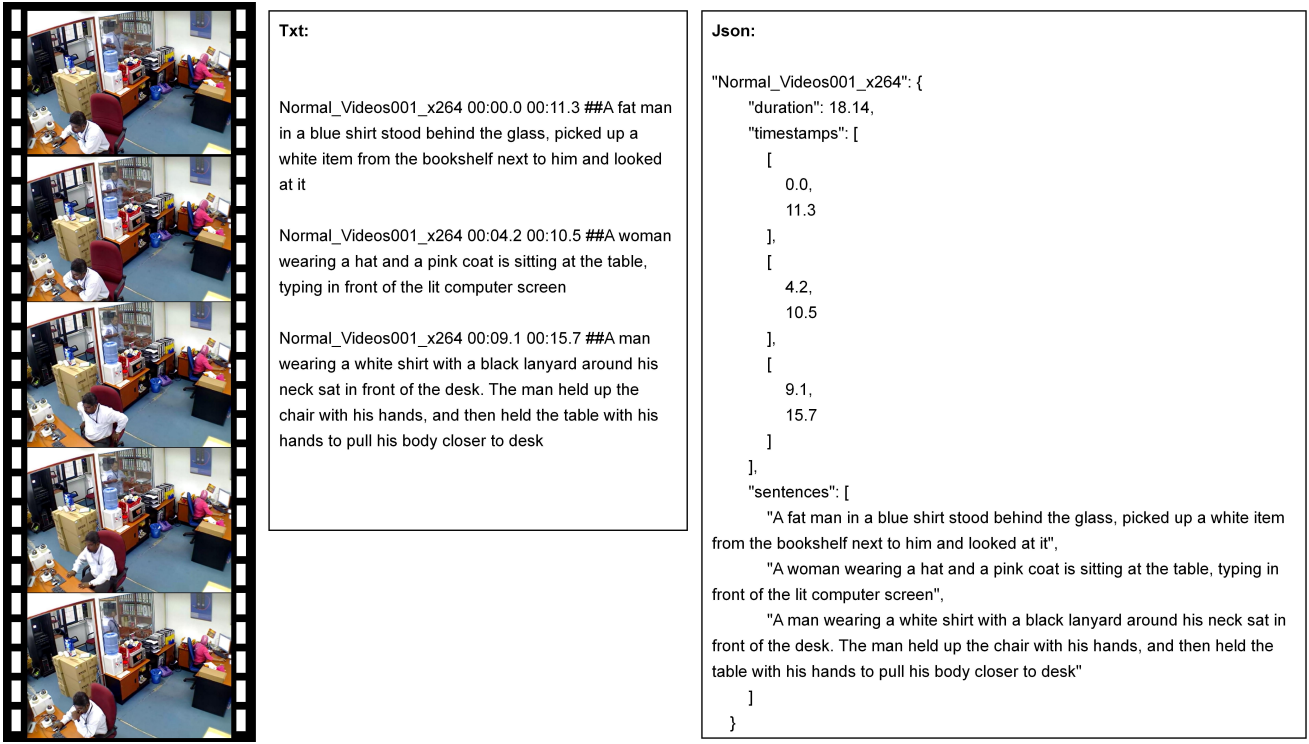}
    \caption{Detailed Example of the Annotation Format.}
    \label{fig-visual}
\end{figure*}
In the txt files, we record several pieces of information for each annotation, including the corresponding video name, event start time, event end time, and annotation content.

In the json files, we record the timestamp and description of each annotation as a list. Additionally, we include the duration information of the original video, which will facilitate subsequent experimental tasks.

Figure \ref{fig-format} illustrates two versions of the annotation format, while Figure \ref{fig-visual} provides a specific example.

\subsection{Annotation Guidelines}

Here are the main annotation guidelines we established prior to commencing the labeling work, aimed at ensuring accuracy and consistency throughout the data annotation process:

\begin{itemize}

\item Fine-grained annotation principle: we emphasize the granularity of the data, considering each event with changes (including changes in human or object states) as an individual data point.

\item Time precision: time should be accurately recorded up to 0.1 seconds, allowing for overlaps in the start and end times of multiple samples.

\item Rich sentence descriptions: we encourage the use of rich sentence descriptions that enhance semantic understanding, employing techniques such as the use of adjectives to provide detailed coverage of the objects present in the scene.

\item Region of Interest (ROI) descriptions: in cases where multiple regions in a single frame experience simultaneous actions or events, annotators may use terms like "top-left corner" or "middle" to differentiate and describe the changes in the states of people or objects within those specific regions.

\item Handling intense changes: For scenes with rapid changes, annotators should combine short-term and long-term descriptions. Initially, they can describe these changes over shorter intervals to capture the micro-variations in the video content. Then, these scattered short-term descriptions should be integrated into a coherent and comprehensive narrative. This layered, multi-granularity annotation method can not only improve the accuracy of the descriptions but also ensure a complete representation of the event, playing a crucial role in video anomaly detection and content analysis.

\item Clear action descriptions: If actions in a video are so subtle that they are hard to discern by the human eye, unclear action descriptions can affect the overall quality of the annotation. Hence, we recommend ensuring clarity and accuracy in action descriptions during annotation. For video segments where actions are not obvious or difficult to discern, we decide whether to include them in the dataset based on their importance. In some cases, if these segments' actions are of little value to the study or their ambiguity could lead to misunderstandings, we might choose to omit these annotations or, in extreme cases, remove these videos from the dataset. Under this condition, we encourage annotators to communicate with our reviewer team.

\item Complex environment descriptions: In complex environments, where there may be a lack of obvious main events or prominent actions, extracting key information from videos becomes challenging. Therefore, we encourage annotators to provide an overall overview to facilitate a comprehensive understanding of the environment.

\end{itemize}

In our annotation project, experts initially annotated the video data, collecting 100 examples. These cases not only laid the groundwork for our preliminary version of the annotation guidelines but also became a resource for subsequent annotators to refer to. Throughout the annotation process, we encourage annotators to communicate with our review team whenever they encounter uncertainties or questions. Through this continuous exchange and feedback, we have constantly refined and fine-tuned these annotation principles. \\
These meticulously designed principles provide clear guidance to annotators throughout the process, ensuring high quality and consistency of the annotated data.

\subsection{Analysis on Annotation Agreement}

Ensuring agreement among different annotators is a critical aspect of the annotation process. During the dataset annotation process, we recruited 10 volunteers with computer backgrounds as annotators and formed a review team consisting of 3 AI researchers. To achieve this, we impose specific constraints and guidelines to maintain a high level of consistency throughout the annotation process. These constraints include:
\begin{itemize}
\item Detailed Annotation Guidelines: we provide annotators with comprehensive guidelines that outline the specific criteria and principles for annotation.

\item Training and Familiarization: before starting the annotation work, annotators undergo training sessions where they are familiarized with the annotation guidelines and are given the opportunity to clarify any doubts or concerns. 

\item Ongoing Communication: throughout the annotation process, annotators are encouraged to engage in regular communication with designated reviewers. This allows them to seek clarification on any ambiguous aspects of the annotation task and receive feedback on their work. 

\item Review and Validation: the work of annotators undergoes thorough review and validation by designated reviewers. This process involves cross-checking annotations against the guidelines and comparing them with the work of other annotators. During the review process, particular attention has been given to addressing disagreement in annotation styles among annotators. Any discrepancies or inconsistencies are identified and addressed through feedback and clarification.
\end{itemize}

By implementing these constraints and measures, we aim to ensure a high level of agreement among annotators, which enhances the reliability and quality of the annotated dataset.

\section{Experiment Details}
To ensure consistency, all experiments are conducted using an RTX3090 GPU with 24GB of memory. The experimental environment is configured with CUDA 11.4, Python 3.9, PyTorch 1.12.0, and TensorFlow 2.12.0. However, due to version compatibility issues in the source code of TDA-CG, a Tesla T4 GPU with 16GB of memory is utilized for running this specific experiment. The environment for this experiment is configured with CUDA 10.1, Python 2.7, and TensorFlow 1.14.0.

In the following, we will present the experimental setup specifics for different experimental tasks. Simultaneously, to compare and demonstrate the efficiency differences among various methods, we also record the runtime of each model.
\subsection{Experimental Details of TSGV}

\begin{figure}[h]
    \centering
    \includegraphics[width=\linewidth]{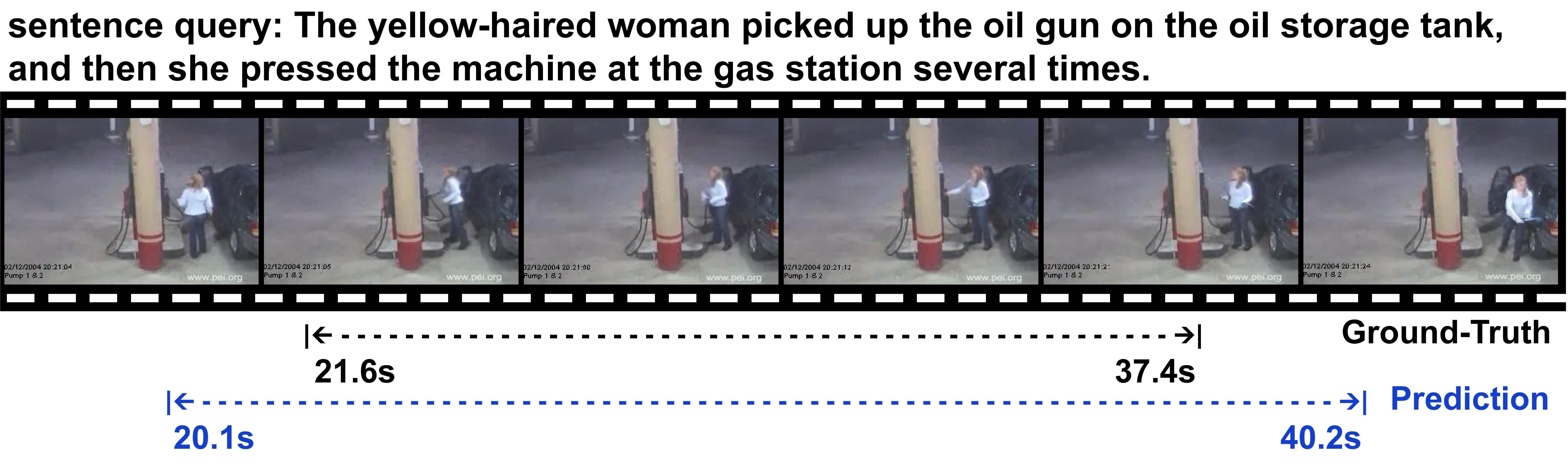}
    \caption{Visualization results of TSGV}
    \label{fig-tsgv-2}
\end{figure}

In the Temporal Sentence Grounding in Videos (TSGV) task, we opt to use the C3D network pre-trained on the Sports1M dataset for extracting visual features. As this network is used for visual feature extraction in all experimental models across various datasets, we chose the C3D network for visual feature extraction on the UCA dataset as well, to facilitate experimentation and enable easy comparison with results from other datasets.

For CTRL~\cite{gao2017tall}, we set the sliding window size to 128 or 240 frames, equivalent to about 5 or 10 seconds of content in the original video, with an overlap of 0.5. The number of context clips is set to 1. The text encoder employs the Skip-thought model, producing text features of 4800 dimensions. During training, the batch size is set to 32, using the Adam optimizer.

For SCDM~\cite{yuan2019semantic}, the input video clip length is set to 512 frames, longer videos are truncated, and shorter ones are padded with zero vectors. The settings for the temporal convolutional layers are five temporal dimensions \{128, 64, 32, 16, 8\}. The maximum length for input text is limited to 50, with word embedding sequences obtained through Glove. During training, the batch size is set to 8, using the Adam optimizer.

For A2C~\cite{he2019read}, when setting normalized start and end points, the maximum frame length  $T_{max}$  is set to 10. In the observation network, the output size of the fully connected layer for encoding description features is set to 1024. The state vector s(t) at step t is also 1024 dimensions. The text encoder uses the Skip-thought model. During training, the batch size is 32, using the Adam optimizer.

For LGI~\cite{mun2020local}, the model uniformly samples 128 segments from each video, with the maximum text length set to 50, and $\lambda$, which controls the extent of overlap between query attention distributions, is set to 0.2. During training, the batch size is 64, using the Adam optimizer.

For 2D-TAN~\cite{zhang2020learning}, the number of sampled clips $N$ in visual features is set to 16. Non-maximum suppression with a threshold of 0.5 is applied during inference. The network structure uses an 8-layer convolutional network with a kernel size of 5. During training, the batch size is 32, using the Adam optimizer.

For MMN~\cite{wang2022negative}, the number of sampled clips $N$ for visual features is also set to 16, the dimension of the joint feature space $d^{H}$ is 256, and the temperature parameter is set to 0.1. During training, the batch size is 8, using the Adam optimizer.

For MomentDiff~\cite{li2023momentdiff}, the maximum length of visual features is set to 1000, the maximum text length to 32, and the number of random spans $N_{r}$ to 5. During training, the batch size is 16, using the Adam optimizer.

In all experiments, the CTRL model exhibits the longest runtime, requiring about 22 hours per training and testing epoch. This is mainly due to the original model processing a large number of clip-sentence pairs. Each training and testing epoch of the A2C model takes about 50 minutes. The epochs for 2D-TAN and SCDM take about 1 hour each. The MomentDiff model requires approximately 20 minutes per epoch. Among all the models, MMN performs the most efficiently, completing each training and testing round within 10 minutes.

\subsection{Experimental Details of Video Captioning}
In the video captioning (VC) task, we utilize the pre-trained model library built on PyTorch for extracting visual features.

For S2VT~\cite{venugopalan2015sequence}, we uniformly sample 80 frames from each video, employing pre-trained Inception V4 and VGG16 BN as the visual encoders. During training, the batch size is set to 32, optimized with the Adam optimizer.

For RecNet~\cite{wang2018reconstruction}, the visual encoder also utilizes Inception V4, sampling 80 frames uniformly from each video. The maximum sentence length is limited to 50, with excess being truncated. In training, both local and global parts have a batch size of 32, using the AMSGrad optimizer.

For MARN~\cite{pei2019memory}, we similarly sample 80 frames from each video, with pre-trained Inception V4 and ResNext101 as visual encoders, subsequently linearly transformed to 512 dimensions. The maximum sentence length is set to 50. Initially, 100 epochs of training are conducted on the Attention-based Recurrent Decoder, followed by the integration of the Attended Memory Decoder. During training, the batch size is 32, using the Adam optimizer.

For SGN~\cite{ryu2021semantic}, 50 frames are uniformly sampled from each video, using pre-trained ResNet 101 and ResNext101 as visual encoders to extract 2D and 3D features. The maximum sentence length is set to 30, with word embedding sequences obtained through GloVe. The model's similarity threshold $\tau$ is set to 0.2, and the coefficient $\lambda$ for Contrastive Attention loss is 0.16. During training, the batch size is 16, using the Adamax optimizer.

For SwinBERT~\cite{lin2022swinbert}, in the preparation stage, each video is segmented into 32 frames, employing an end-to-end training approach. VidSwin is initialized with pre-trained weights from Kinetics-600, and the multimodal transformer encoder is randomly initialized. Due to GPU memory constraints, the batch size during training is set to 4, using the Adam optimizer.

For CoCap~\cite{shen2023accurate}, before training and testing, the video's minimum edge size is adjusted to 240, and it is compressed using H.264 encoding. The visual encoder is initialized with pre-trained weights from CLIP, while the other encoders and the multimodal decoder are randomly initialized. During training, the batch size is 4, using the Adam optimizer.

In all experiments, the SwinBERT model, due to its end-to-end training mechanism, requires the longest time per epoch for training and testing, approximately 80 minutes. The CoCap model significantly improves training efficiency by using compressed videos, taking about 40 minutes per epoch. The MARN model, after the integration of the Memory Decoder in the training phase, requires about 60 minutes per epoch. The remaining models complete each training and testing epoch in under 20 minutes.

\begin{figure}[ht]
    \centering
    \includegraphics[width=\linewidth]{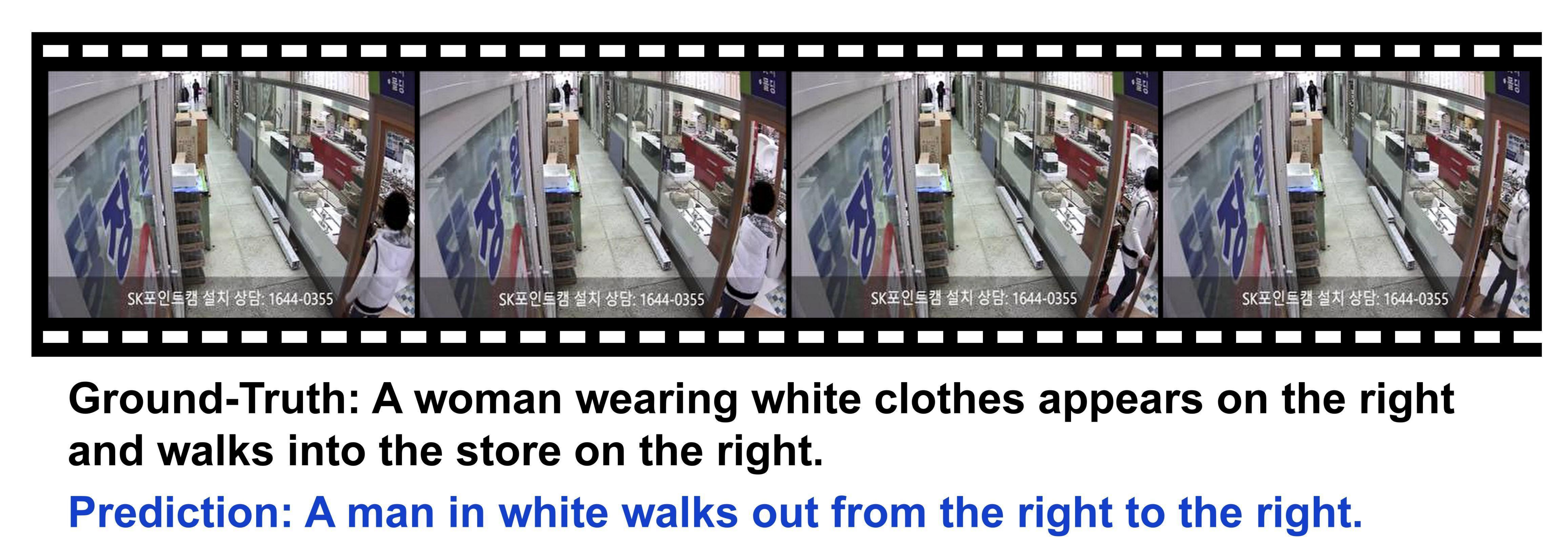}
    \caption{Visualization results of Successful Video Captioning}
    \label{fig-vc2}
\end{figure}

\begin{figure}[ht]
    \centering
    \includegraphics[width=\linewidth]{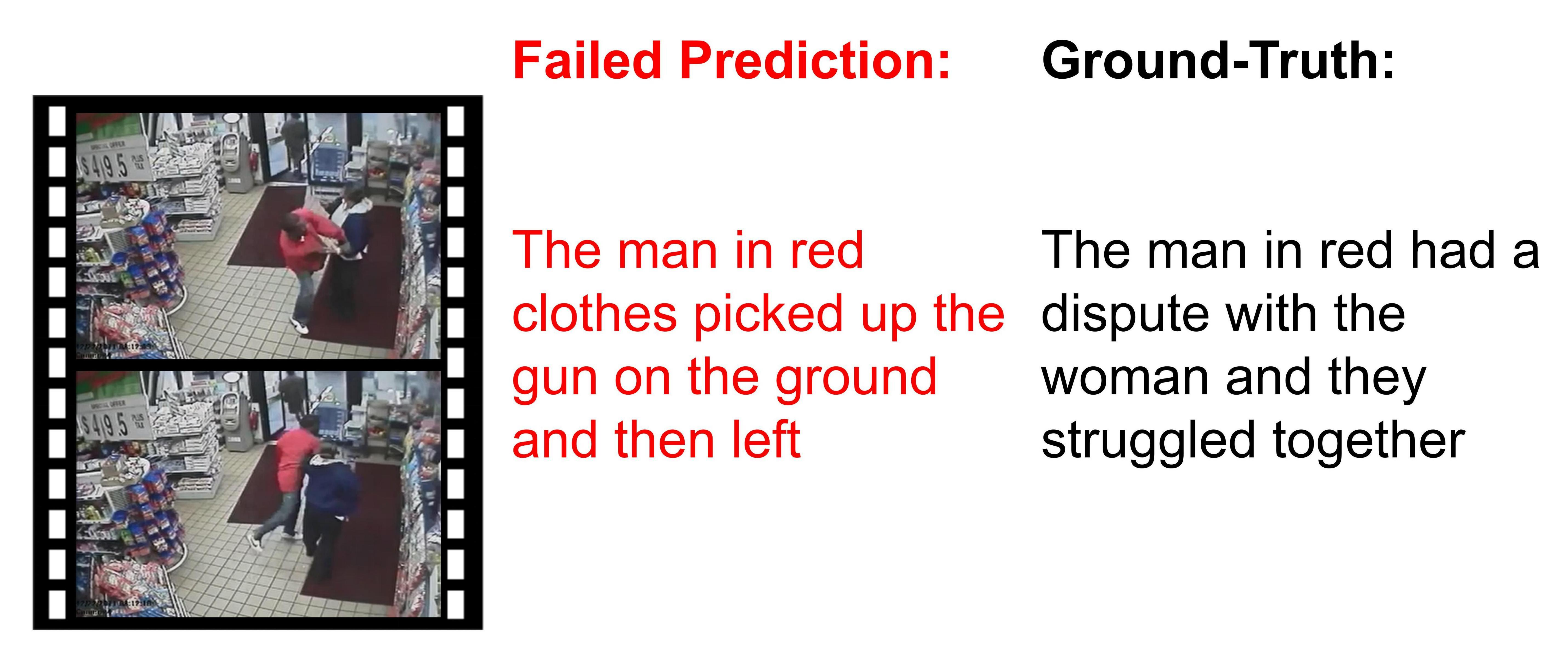}
    \caption{Visualization results of Failed Video Captioning}
    \label{fig-vc-failed}
\end{figure}

\subsection{Experimental Details of Dense Video Captioning}

\begin{figure}[ht]
    \centering
    \includegraphics[width=\linewidth]{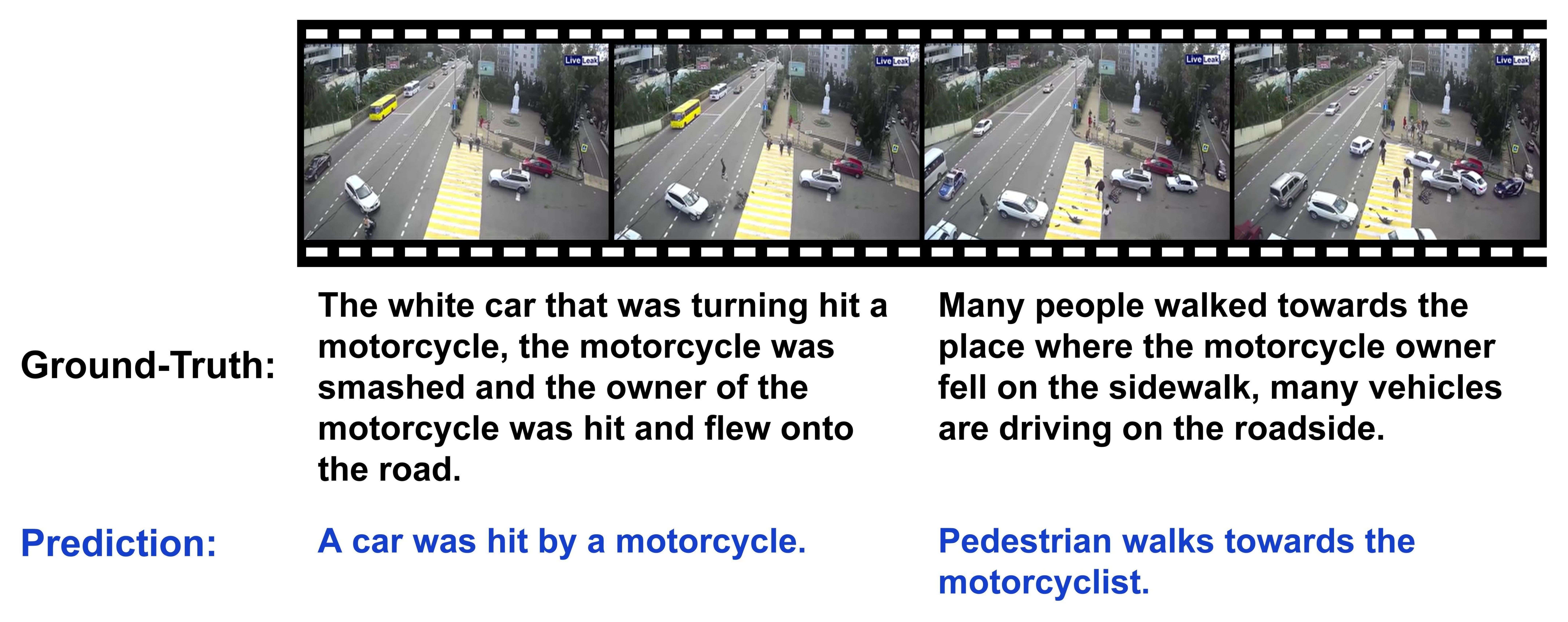}
    \caption{Visualization results of Successful Dense Video Captioning}
    \label{fig-dvc}
\end{figure}
\begin{figure}[ht]
    \centering
    \includegraphics[width=\linewidth]{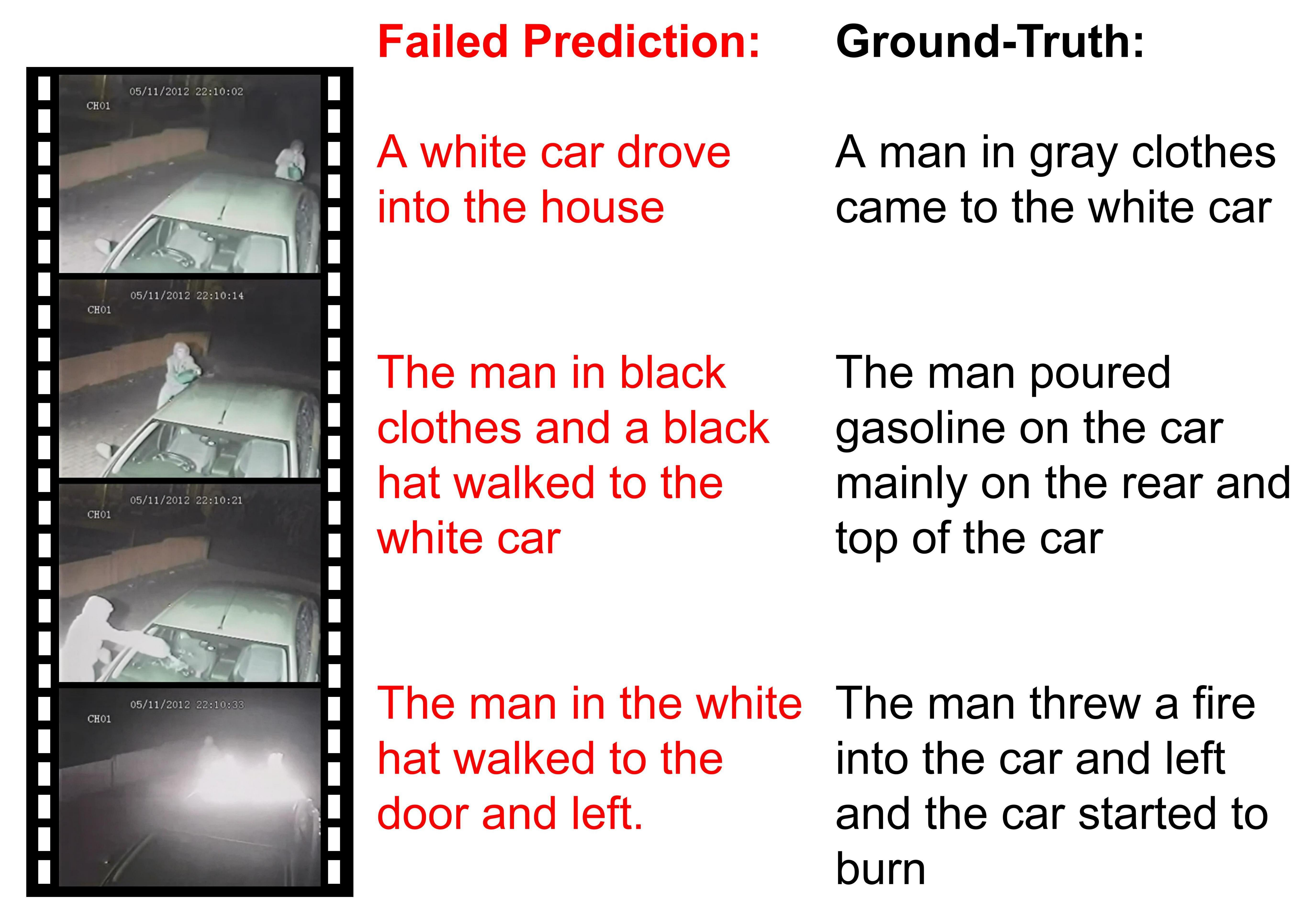}
    \caption{Visualization results of Failed Dense Video Captioning}
    \label{fig-dvc-failed}
\end{figure}

In this dense video captioning (DVC) task, the C3D features used remain consistent with those used in TSGV. The I3D features are directly obtained from an external GitHub repository at \url{https://github.com/tianyu0207/RTFM}, which was established by other researchers. 

For TDA-CG~\cite{wang2018bidirectional}, each feature sequence corresponds to 64 frames of content in the video. The maximum sentence length is limited to 30. During training, the Adam optimizer is used.

For PDVC~\cite{wang2021end}, we set the number of event queries to 100, and conduct experiments using the standard PDVC model with the LSTMDSA captioner. The LSTM hidden layer dimension in the caption head is set to 512. The Adam optimizer is used during training.

For UEDVC~\cite{zhang2022unifying}, the maximum length of video frames is set to 200, and the maximum sentence length is 50. The number of layers in the independent encoder is set to 1, while the cross encoder has 4 layers. This model's training also employs the Adam optimizer.
\subsection{Experimental Details of MAD}
\begin{figure}[h]
    \centering
    \includegraphics[width=0.9\linewidth]{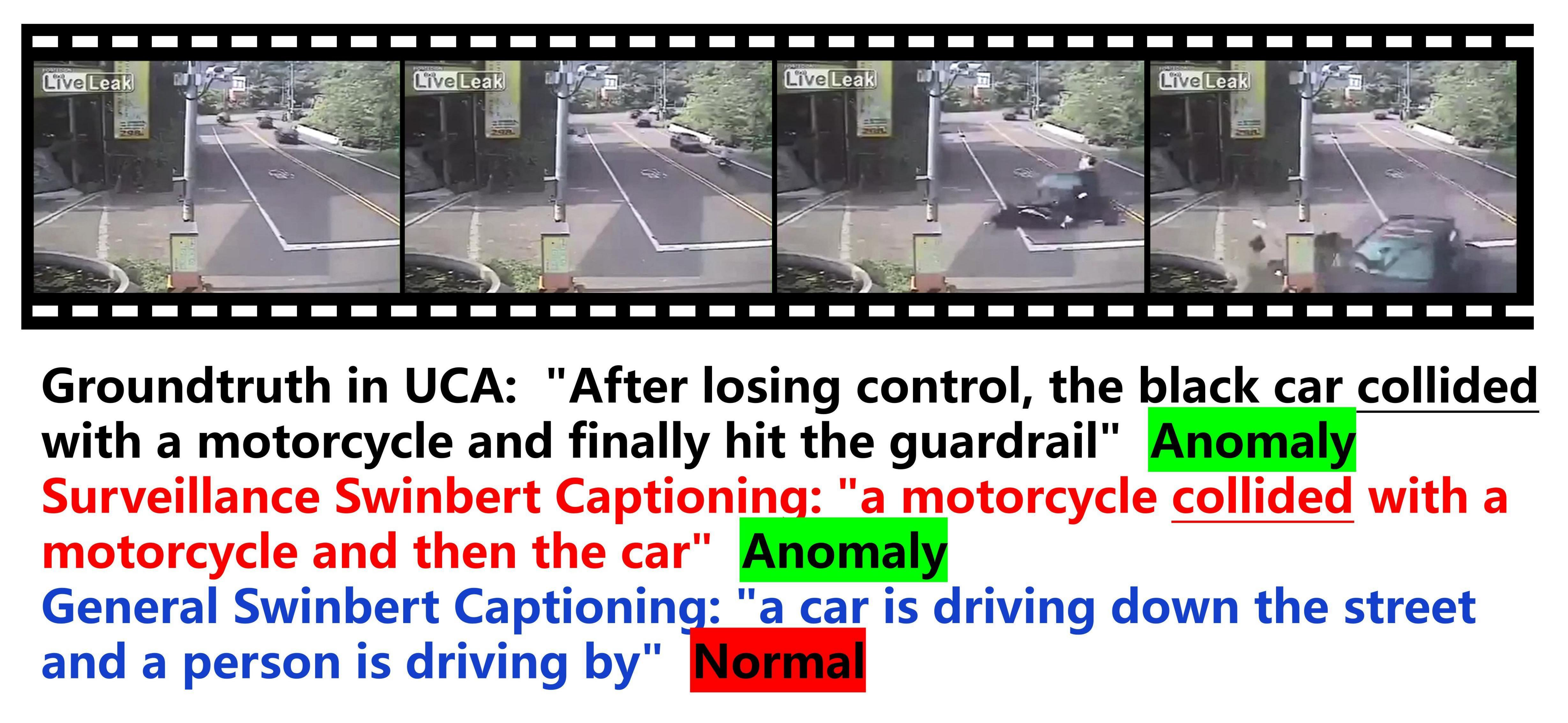}
    \vspace{-1em}
    \caption{Examples of different video captioning results in MAD.}
    \label{fig-MAD}
\end{figure}

\begin{figure}[h]
  \centering

  \begin{subfigure}[b]{\textwidth}
    \includegraphics[width=0.5\textwidth]{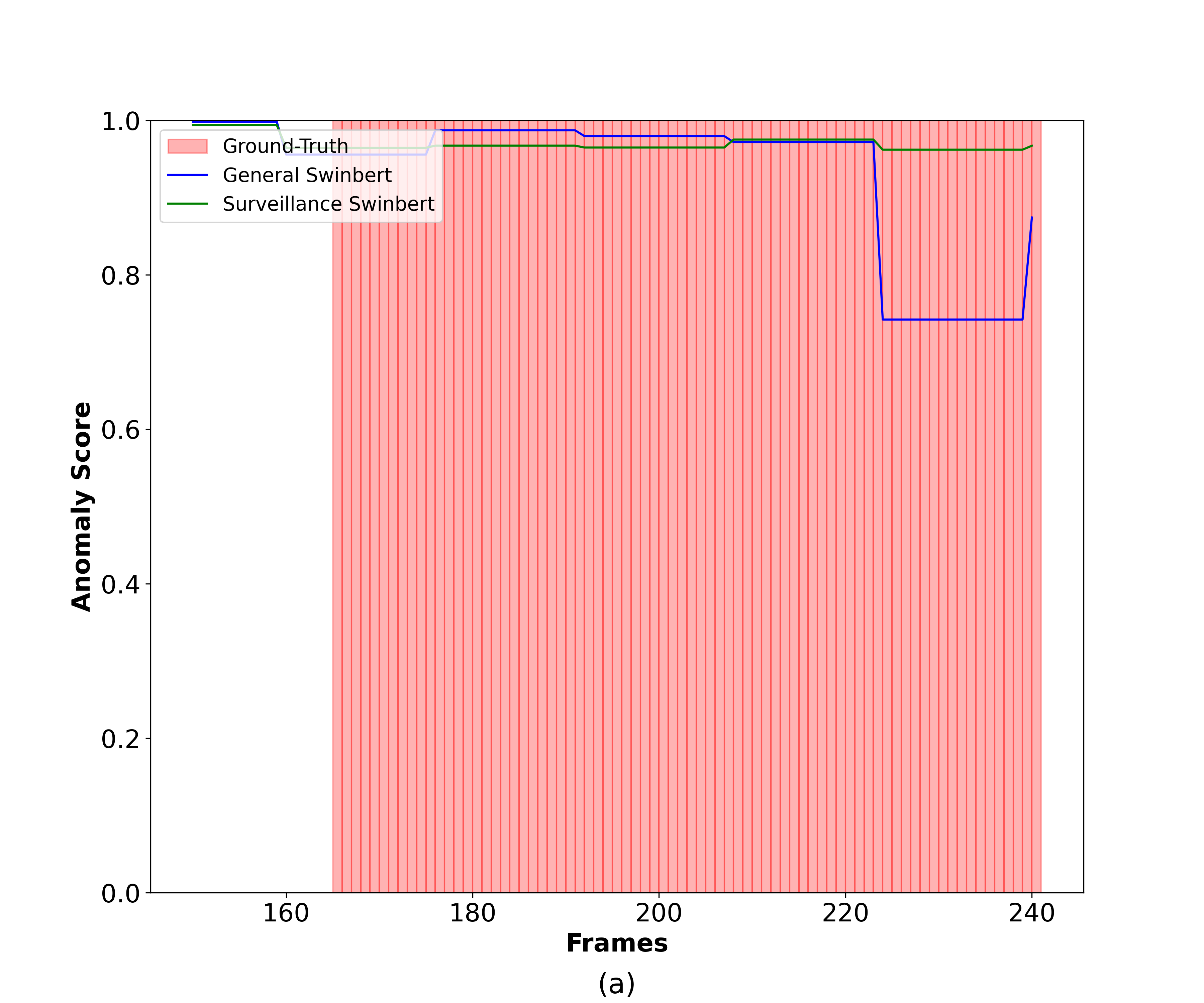}
    \label{fig:sub1}
  \end{subfigure}

  \begin{subfigure}[b]{\textwidth}
    \includegraphics[width=0.5\textwidth]{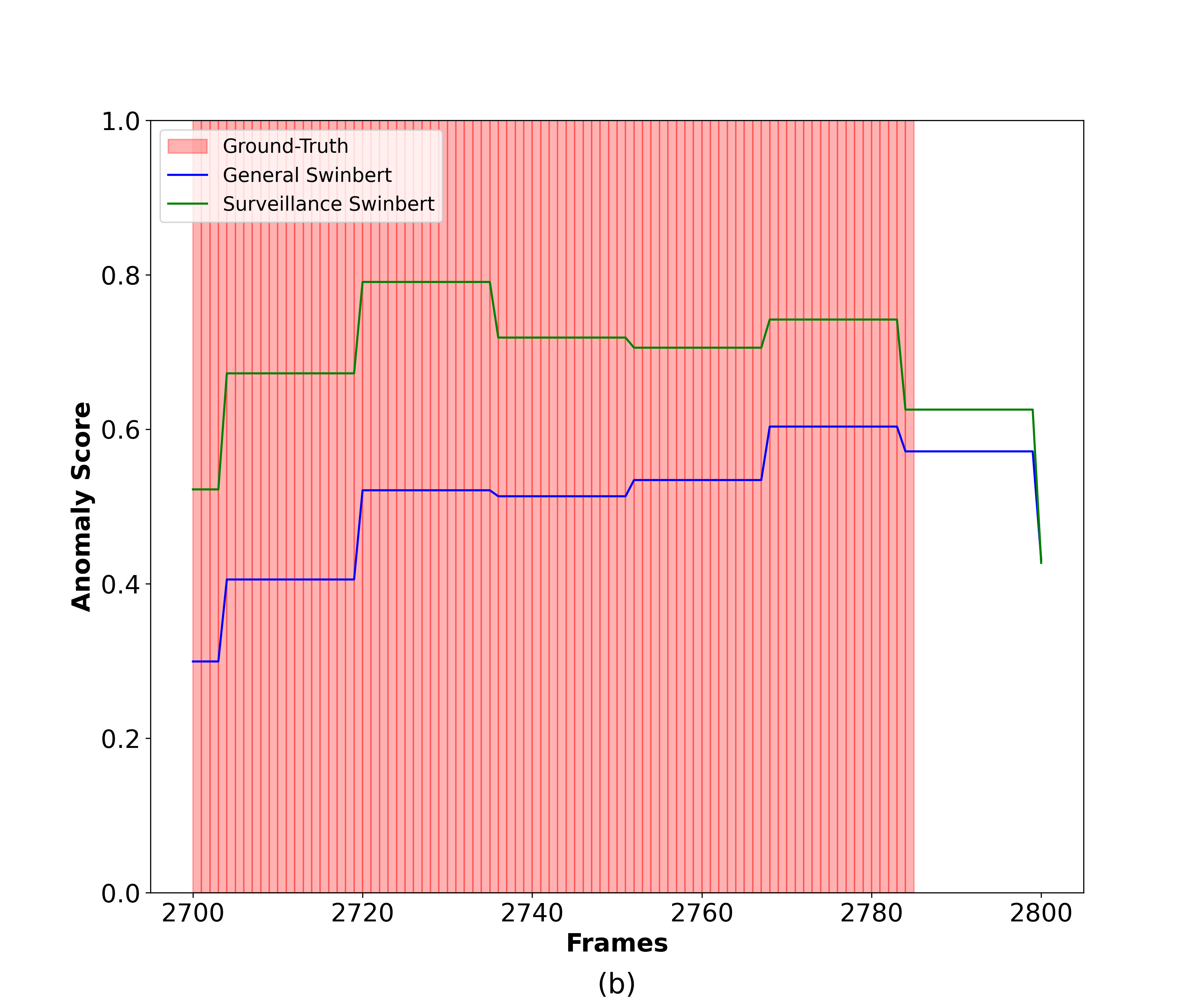}
    \label{fig:sub2}
  \end{subfigure}
  \caption{Comparison of MAD results using different text branches. The magnitude of curves represents the anomaly score. }
  \label{fig-madtest}
\end{figure}
We will introduce the experimental details of the Multimodal Anomaly Detection (MAD) task. 
\\
For TEVAD~\cite{chen2023tevad}, we use the same visual features as the original model. The video is divided into non-overlapping segments of 16 frames, and 2048-dimensional visual features are obtained through the I3D feature extractor. The hyperparameter $\lambda$, used to adjust the weights of the loss components, is set to 0.0001. During training, the batch size is 32, using the Adam optimizer.
\subsection{Visualization Results}

This section presents visualizations of the experimental outcomes outlined within the paper.  Figure \ref{fig-tsgv-2} showcases instances where the IoU is greater than 0.7 between predicted and ground truth results in the TSGV task.   Figures \ref{fig-vc2} and \ref{fig-vc-failed} respectively depict accurate predictions and erroneous predictions in the Video Captioning task.  Furthermore, Figures \ref{fig-dvc} and \ref{fig-dvc-failed} display successful and unsuccessful predictions in the Video Dense Captioning task.  By visualizing and analyzing these instances of failure, we identify challenges that existing models may encounter when handling surveillance videos, including inaccuracies in color recognition, difficulty in identifying intricate scenes, and struggles in capturing subtle movements. 

Figure \ref{fig-MAD} shows the visualization results in the MAD task. From Figure \ref{fig-MAD}, we can find Surveillance SwinBERT can generate anomaly captions for videos, which is more similar to ground truth in UCA. However, General SwinBERT generates normal captions, which are different from ground truth in UCA. Figure \ref{fig-madtest} illustrates the impact of different text branches on anomaly detection result scores in MAD. The two subplots extract frames 150-240 and 2,700-2,800 from the video in UCF-Crime, with the red region indicating frames labeled as anomalies in the ground truth. The two curves represent the results of different text branches obtained using General SwinBERT and Surveillance SwinBERT. Clearly, when using Surveillance SwinBERT to generate descriptive statements, anomalous video frames obtain higher anomaly scores, thereby improving the accuracy of anomaly detection. This further demonstrates the effectiveness of introducing the UCA dataset for the anomaly detection task.

These findings contribute to the exploration of the potential value of the new UCA dataset in enhancing anomaly detection model performance.

\section{Additional Experiments}

\subsection{Multimodal Task Performance on Normal and Anomalous videos}
The video data in UCF-Crime can be classified into two major categories: abnormal videos and normal videos. To delve into the performance differences between these two types of videos, we conduct a series of tests using pre-trained models, evaluating the best-performing methods for each task.

In the TSGV task, as shown in Table~\ref{normal-TSGV}, the experimental results indicate that normal videos significantly underperform compared to abnormal videos. This finding aligns with our expectations. The TSGV task focuses on locating specific video segments within untrimmed videos. Given that abnormal videos contain more distinctive segments that starkly contrast with regular scenes, they are easier to identify and locate. We also analyzed the differences in video length between abnormal and normal videos, finding that normal videos contain more long-duration videos, which is a primary reason for the difficulty in event localization in these videos.

In the VC task,  the experimental results are presented in Table~\ref{normal-vc}. This task requires generating textual descriptions for video segments, where the original length of the video does not influence the outcome. Abnormal videos often contain specific descriptive terms like 'explosion,' 'collision,' etc., which are less frequent in the vocabulary. Precisely describing anomalous events in surveillance videos remains a challenge for existing models, hence normal videos with more generic descriptions perform better in tests.

In the DVC task, as shown in Table~\ref{normal-dvc}, the results suggest that normal videos outperform abnormal videos in terms of subtitle generation accuracy. However, since this task requires complete video features as input, the difference in video length between normal and abnormal videos imposes limitations on the models in capturing key video information and narrating the video story.

\begin{table}[t]
\centering
\caption{Performance of TSGV on UCA Dataset}
\label{normal-TSGV}
\resizebox{0.9\linewidth}{!}{%
\begin{tabular}{lccccccc}
\toprule
\multirow{2}{*}{Method}  & \multirow{2}{*}{Split}& \multicolumn{2}{c}{IoU=0.3}      & \multicolumn{2}{c}{IoU=0.5}      & \multicolumn{2}{c}{IoU=0.7}      \\ \cmidrule{3-8}
                                                        &         & R@1 & R@5   & R@1  & R@5   &R@1  & R@5    \\ \midrule

\multirow{2}{*}{MMN ~\cite{wang2022negative}}                          & \multicolumn{1}{c}{Normal}   & \multicolumn{1}{c}{4.02} & 10.73 & \multicolumn{1}{c}{2.14} & 5.63 & \multicolumn{1}{c}{1.09} & 3.11        \\ 
 & \multicolumn{1}{c}{Anomalous}   & \multicolumn{1}{c}{16.15} & 38.48 & \multicolumn{1}{c}{8.70} & 22.15 & \multicolumn{1}{c}{3.78} & 10.32        \\ 

\bottomrule
\end{tabular}}%
\end{table}

\begin{table}[t]
\centering
\caption{Performance of VC on UCA Dataset.}
\label{normal-vc}
\resizebox{1.\linewidth}{!}{%
\begin{tabular}{llccccccccc}
\toprule
Method         & Split                                            & B1    & B2    & B3   & B4   & M    & R     & C     \\ \midrule
\multirow{2}{*}{CoCap~\cite{shen2023accurate}}      & Normal       &27.55  &17.27   &11.04  &6.98  &11.79 &29.41   &21.75            \\
& Anomalous                     &29.55  &16.46   &9.2  &4.82 &10.57 & 26.10  &18.62        \\
\bottomrule
\end{tabular}}%
\end{table}

\begin{table}[t]
    \centering
    \caption{Performance of DVC on UCA Dataset.}
    \vspace{-1em}
    \label{normal-dvc}
\resizebox{1.\linewidth}{!}{
\begin{tabular}{ccccccccc}
\hline
\multicolumn{1}{c}{\multirow{2}{*}{Method}} & \multicolumn{1}{c}{\multirow{2}{*}{Split}} & \multicolumn{7}{c}{Predicted proposals}                                                                                                                                                            \\ \cline{3-9} 
\multicolumn{1}{c}{}                        & \multicolumn{1}{c}{}                          & \multicolumn{1}{c}{B1}   & \multicolumn{1}{c}{B2}   & \multicolumn{1}{c}{B3}   & \multicolumn{1}{c}{B4}   & \multicolumn{1}{c}{M}    & \multicolumn{1}{c}{C}    & \multicolumn{1}{l}{SODA\_c} \\ \hline
\multicolumn{1}{c}{\multirow{2}{*}{PDVC~\cite{wang2021end}}}                   & \multicolumn{1}{c}{Normal}                       & \multicolumn{1}{c}{8.94} & \multicolumn{1}{c}{4.97} & \multicolumn{1}{c}{2.66} & \multicolumn{1}{c}{1.12} & \multicolumn{1}{c}{4.21} & \multicolumn{1}{c}{7.88} & \multicolumn{1}{c}{1.80}   \\ 
 & \multicolumn{1}{c}{Anomalous}                       & \multicolumn{1}{c}{7.56} & \multicolumn{1}{c}{3.85} & \multicolumn{1}{c}{1.67} & \multicolumn{1}{c}{0.50} & \multicolumn{1}{c}{3.99} & \multicolumn{1}{c}{9.23} & \multicolumn{1}{c}{2.69}   \\
\hline

\multicolumn{1}{c}{\multirow{2}{*}{Method}} & \multicolumn{1}{c}{\multirow{2}{*}{Split}} & \multicolumn{7}{c}{Ground-Truth proposals}                                                                                                                                                            \\ \cline{3-9} 
\multicolumn{1}{c}{}                        & \multicolumn{1}{c}{}                          & \multicolumn{1}{c}{B1}   & \multicolumn{1}{c}{B2}   & \multicolumn{1}{c}{B3}   & \multicolumn{1}{c}{B4}   & \multicolumn{1}{c}{M}    & \multicolumn{1}{c}{C}    &  \\ \hline
\multicolumn{1}{c}{\multirow{2}{*}{PDVC~\cite{wang2021end}}}                   & \multicolumn{1}{c}{Normal}                       & \multicolumn{1}{c}{22.16} & \multicolumn{1}{c}{12.73} & \multicolumn{1}{c}{7.15} & \multicolumn{1}{c}{3.61} & \multicolumn{1}{c}{10.32} & \multicolumn{1}{c}{17.44} &    \\ 
 & \multicolumn{1}{c}{Anomalous}                       & \multicolumn{1}{c}{24.33} & \multicolumn{1}{c}{12.48} & \multicolumn{1}{c}{5.28} & \multicolumn{1}{c}{1.97} & \multicolumn{1}{c}{10.68} & \multicolumn{1}{c}{25.26} &    \\
\hline
\end{tabular}
  }
\end{table}

\subsection{Multimodal Task Performance with Gender-Neutral Annotations}
To minimize the impact of gender on experimental results, we generated gender-neutral annotations, as detailed in Table~\ref{vocabulary}. For three distinct multimodal tasks, we selected one model for experimentation using gender-neutral annotations.

The experimental results for the TSGV task are presented in Table~\ref{gender-TSGV}. It can be observed that, at low precision (IoU=0.3), gender-neutral annotations show a slight improvement compared to the regular version. However, as the task difficulty increases, there is no significant difference between the results of gender-neutral and regular versions. This further underscores the challenge of accurately pinpointing "what a person is doing" on the UCA dataset.

The experimental results for the VC and DVC tasks are presented in Table ~\ref{gender-vc} and Table ~\ref{gender-dvc}. Both tasks require generating captions based on video content. Substituting different gender-specific terms with neutral terms in the vocabulary contributes to a more uniform vocabulary, leading to slightly higher experimental results using gender-neutral annotations compared to the regular version.

Overall, the impact of using gender-neutral annotations on these multimodal tasks is relatively minor. As mentioned in the paper, different tasks still face significant challenges in the surveillance video dataset. We provide gender-neutral annotations in the repository, and researchers considering ethical considerations are encouraged to utilize this version of annotations.
\begin{table}[]
\centering
\caption{Replacement vocabulary for gender-neutral annotations.}
\label{vocabulary}
\resizebox{.9\linewidth}{!}{%
\begin{tabular}{cc}
\hline
\multicolumn{1}{c}{Regular} & \multicolumn{1}{c}{Gender-neutral} \\ \hline
woman, man, she, he, him          & person                              \\
herself, himself               & themself                            \\
her, his                       & the person's                        \\
policeman                     & police                              \\
salesman                      & salesperson                         \\
postman                       & mail carrier                        \\
doorman                       & doorperson                          \\
fireman                       & firefighter                         \\
gunman                        & person with a gun                   \\
repairman                     & mechanic                            \\
cameraman                     & photographer        \\ \hline               
\end{tabular}}
\end{table}

\begin{table}[t]
\centering
\caption{Performance of TSGV on UCA Dataset with Gender-Neutral Annotations.}
\label{gender-TSGV}
\resizebox{0.9\linewidth}{!}{%
\begin{tabular}{lcccccc}
\toprule
\multirow{2}{*}{Method}  & \multicolumn{2}{c}{IoU=0.3}      & \multicolumn{2}{c}{IoU=0.5}      & \multicolumn{2}{c}{IoU=0.7}      \\ \cmidrule{2-7}
                                                                                     & R@1 & R@5   & R@1  & R@5   &R@1  & R@5    \\ \midrule

MMN ~\cite{wang2022negative}                                                            & \multicolumn{1}{c}{9.03} & 21.77 & \multicolumn{1}{c}{4.13} & 12.42 & \multicolumn{1}{c}{2.08} & 5.82        \\ 

\bottomrule
\end{tabular}}%
\end{table}

\begin{table}[ht]
\centering
\caption{Performance of VC on UCA Dataset with Gender-Neutral Annotations.}
\label{gender-vc}
\resizebox{1.\linewidth}{!}{%
\begin{tabular}{llccccccccc}
\toprule
Method         & Features                                             & B1    & B2    & B3   & B4   & M    & R     & C     \\ \midrule
CoCap~\cite{shen2023accurate}      & CLIP                                                                &30.04  &18.71   &11.58  &6.89  &12.22&29.48  &20.41           \\
\bottomrule
\end{tabular}}%
\end{table}

\begin{table}[h]
    \centering
    \caption{Performance of DVC on UCA Dataset with Gender-Neutral Annotations.}
    \vspace{-1em}
    \label{gender-dvc}
\resizebox{1.\linewidth}{!}{
\begin{tabular}{ccccccccc}
\hline
\multicolumn{1}{c}{\multirow{2}{*}{Method}} & \multicolumn{1}{c}{\multirow{2}{*}{Features}} & \multicolumn{7}{c}{Predicted proposals}                                                                                                                                                            \\ \cline{3-9} 
\multicolumn{1}{c}{}                        & \multicolumn{1}{c}{}                          & \multicolumn{1}{c}{B1}   & \multicolumn{1}{c}{B2}   & \multicolumn{1}{c}{B3}   & \multicolumn{1}{c}{B4}   & \multicolumn{1}{c}{M}    & \multicolumn{1}{c}{C}    & \multicolumn{1}{l}{SODA\_c} \\ \hline
\multicolumn{1}{c}{PDVC~\cite{wang2021end}}                   & \multicolumn{1}{c}{I3D}                       & \multicolumn{1}{c}{9.08} & \multicolumn{1}{c}{5.13} & \multicolumn{1}{c}{2.71} & \multicolumn{1}{c}{1.04} & \multicolumn{1}{c}{4.72} & \multicolumn{1}{c}{10.06} & \multicolumn{1}{c}{2.50}   \\ \hline
\multicolumn{1}{c}{\multirow{2}{*}{Method}} & \multicolumn{1}{c}{\multirow{2}{*}{Features}} &\multicolumn{7}{c}{Ground-Truth proposals}                                                                                                                                                            \\ \cline{3-9} 
\multicolumn{1}{c}{}                        & \multicolumn{1}{c}{}                          & \multicolumn{1}{c}{B1}   & \multicolumn{1}{c}{B2}   & \multicolumn{1}{c}{B3}   & \multicolumn{1}{c}{B4}   & \multicolumn{1}{c}{M}    & \multicolumn{1}{c}{C}     \\ \hline
\multicolumn{1}{c}{PDVC~\cite{wang2021end}}                   & \multicolumn{1}{c}{I3D}                       & \multicolumn{1}{c}{25.18} & \multicolumn{1}{c}{14.05} & \multicolumn{1}{c}{7.12} & \multicolumn{1}{c}{2.84} & \multicolumn{1}{c}{11.58} & \multicolumn{1}{c}{23.06} &   \\ \hline
\end{tabular}
  }
\end{table}

\subsection{Anomaly detection}
\begin{table}[h]
\centering
\vspace{-0.5em}
\caption{Comparative Results of Anomaly Detection Accuracy Using Multimodal and Visual-Only Features. Visual* indicates results obtained from models retrained with unified visual features. Multimodal represents using our provided SwinBERT trained on the UCA dataset.}
\vspace{-0.5em}
\label{vad}
\resizebox{0.8\linewidth}{!}{
\begin{tabular}{cccc}
\toprule
Method & Visual* & Multimodal & AUC \\
\midrule
\multirow{2}{*}{MGFN~\cite{chen2023mgfn}} &\checkmark & \ding{55} & 82.42\% \\
              &\checkmark & \checkmark & 83.06\% \\
\hline
\multirow{2}{*}{UR-DMU~\cite{zhou2023dual}} &\checkmark & \ding{55} & 83.14\% \\
              &\checkmark & \checkmark & 84.16\% \\
 \bottomrule
\end{tabular}
}
\end{table}

In this section, we conduct additional anomaly detection experiments, aimed at validating the effectiveness of textual information in enhancing anomaly detection results. We choose two models, MGFN~\cite{chen2023mgfn} and UR-DMU~\cite{zhou2023dual}, and perform experiments under the environment of Python 3.9, CUDA 11.6, and PyTorch 1.13. For comparison convenience, we adopt the same 2048-dimensional I3D visual features as in TEVAD. Moreover, we generate caption texts using the SwinBERT model trained on the UCA dataset, then obtain 768-dimensional sentence embeddings through the supervised SimCSE pretrained on bert-base-uncased. Finally, we concatenate the visual and textual features to serve as the multimodal input features for the models.

For MGFN, we set the random seed to 2023 and retrain the model with both single visual features and multimodal features. During training, the batch size is set to 16 and the learning rate to 0.001. For UR-DMU, the input dimension of visual features is set to 2048, different from the 1024 dimensions used in the original paper. In training, the batch size is set to 32 and the learning rate to 0.0001. Other parameter settings remain the same as in the original paper.

The experimental results are shown in Table \ref{vad}. It is evident that using multimodal features effectively enhances the accuracy of anomaly detection in surveillance videos. Notably, the results of the experiments with the single visual feature branch are obtained by retraining the model, differing from the data provided in the original paper. This part of the experiments aims to highlight the important role of multimodal information in anomaly detection, and the results convincingly demonstrate the effectiveness and significance of our UCA dataset in improving the accuracy of anomaly detection tasks.

\section{License Details and Accessibility}
The UCA dataset is released under Apache License 2.0. In accordance with the Apache License 2.0, users are free to use, modify, and distribute this dataset, but must include the original copyright and license notices. This means that any derivative works or distributed versions based on this dataset should retain the original copyright and license information.

Please note that our dataset is intended solely for academic and research purposes. We encourage the academic community and researchers to use this dataset to advance the field. If you have any questions about the use of the dataset, please contact us directly. We warmly welcome and look forward to your feedback and usage experiences.

The content of the UCA dataset can be accessed at the following link: \url{https://xuange923.github.io/Surveillance-Video-Understanding}. We are committed to providing accessible and user-friendly resources to contribute to the advancement of the field of multimodal surveillance video datasets. We sincerely hope this dataset becomes a valuable resource in your research endeavors.

\end{document}